\documentclass[runningheads]{llncs}
\usepackage[T1]{fontenc}
\usepackage{graphicx,verbatim}
\usepackage{amsfonts,amssymb, amsmath}
\usepackage{booktabs} 
\usepackage{pifont}
\usepackage{makecell}
\usepackage{multirow}
\usepackage[capitalise]{cleveref}
\usepackage[acronym,shortcuts]{glossaries}
\newcommand{\cmark}{\ding{51}}
\usepackage{enumitem}
\newcolumntype{C}[1]{>{\centering\arraybackslash}m{#1}}

\newacronym{ema}{EMA}{exponential moving average}
\begin{document}
%
%\title{Leveraging Unpaired TVUS Prototype Priors for Cross-Modal MRI Ovary Segmentation in Endometriosis}
\title{Cross-Modal MRI Ovary Segmentation in Endometriosis Using Unpaired TVUS Prototype Priors}

\titlerunning{Cross-Modal MRI Ovary Segmentation}
% If the paper title is too long for the running head, you can set
% an abbreviated paper title here
%

% \author{Anonymized Authors}  %% Added for anonymized MICCAI submission
% \authorrunning{Anonymized Author et al.}
% \institute{Anonymized Affiliations \\
%     \email{email@anonymized.com}}
\author{Xingjian Kang\inst{1}$^{,\dagger}$
% \orcidID{0009-0005-0590-1734}
%index{Kang, Xingjian}
\and
Lina Felsner\inst{2}
% \orcidID{0000-0001-7695-2612}
%index{Felsner, Lina}
\and
Dominik Perrin\inst{1}
% \orcidID{0009-0009-8651-155X}
%index{Perrin, Dominik}
\and
Daiqi Liu\inst{3}
% \orcidID{0009-0001-0905-3833}
%index{Liu, Daiqi}
\and
Jasmin Arjomandi\inst{4}
% \orcidID{0009-0008-6801-8676}
\and
Franziska Mathis-Ullrich\inst{4}
% \orcidID{0000-0001-5239-5305}
\and
%index{Mathis-Ullrich, Franziska}
Alexandra Stoll\inst{1,4}
% \orcidID{0000-0002-5105-9534}
%index{Stoll, Alexandra}
\and
Katharina Breininger\inst{1}
% \orcidID{0000-0001-7600-5869}
}
%index{Breininger, Katharina}

\institute{Center for AI and Data Science (CAIDAS), Julius-Maximilians-Universität Würzburg, Würzburg, Germany\\
\and
Institute for Computational Imaging and AI in Medicine (CompAI), Technical University of Munich, Munich, Germany\\
\and
Pattern Recognition Lab, Friedrich-Alexander Universität Erlangen-Nürnberg, Erlangen, Germany\\
\and
Department Artificial Intelligence in Biomedical Engineering (AIBE), Friedrich-Alexander-Universität Erlangen-Nürnberg, Erlangen, Germany\\
\email{xingjian.kang@uni-wuerzburg.de}\\
}
\authorrunning{X.~Kang et al.}

\maketitle              % typeset the header of the contribution
%
% \begingroup
% \renewcommand\thefootnote{}\footnotetext{$^{\dagger}$ Corresponding author}
% \addtocounter{footnote}{-1}
% \endgroup
\begin{abstract}
Transvaginal ultrasound (TVUS) and magnetic resonance imaging (MRI) provide complementary information for endometriosis image analysis, yet existing studies mainly focus on single-modality analysis or disease classification, leaving cross-modal ovarian segmentation largely unexplored. In this work, to tackle the increased difficulty of ovary segmentation in MRI due to ovaries' small target size and ambiguous boundaries with surrounding pelvic structures, we propose a dual branch framework for ovary segmentation across TVUS and MRI.
More specifically, by adapting MedSAM3 with TVUS-derived prototype bank, we aim to align anatomically consistent feature representations across both modalities. Extensive experiments are conducted on endometriosis-related TVUS and MRI datasets. 
We observe quantitative and qualitative improvements of over 5 percentage points for the proposed dual-branch approach compared with multiple state-of-the-art methods. Furthermore, our ablation study shows the contribution of individual components such as the prototype bank and the importance of warm-up pretraining in the source TVUS domain.
% we propose a novel framework for ovarian segmentation across TVUS and MRI by adapting MedSAM3, a promptable concept segmentation foundation model, to endometriosis-related imaging data. Specifically, we employ Parameter-Efficient Fine-Tuning (PEFT) to efficiently transfer the model’s prior knowledge to target modality. Furthermore, considering the increased difficulty of ovarian segmentation in MRI due to the small target size and ambiguous boundaries with surrounding pelvic structures, we introduce a prototype-prior contrastive learning strategy to encourage anatomically consistent feature representations across both modalities. Extensive experiments are conducted on endometriosis-related TVUS and MRI datasets to systematically evaluate the proposed framework. 

% Authors must provide keywords and are not allowed to remove this Keyword section.

\end{abstract}
\section{Introduction}
Endometriosis is a chronic gynecological disease characterized by the presence of endometrial-like tissue outside the uterus~\cite{olive2001treatment}. Among its manifestations, ovarian endometriomas (commonly known as chocolate cysts) can develop within or adjacent to the ovaries~\cite{galczynski2019ovarian}. Transvaginal ultrasound (TVUS) and Magnetic Resonance Imaging (MRI) are two widely-used complementary non-invasive imaging modalities to detect and assess such lesions in patients with endometriosis~\cite{daniilidis2022transvaginal,thomassin2025esur}. 
To support lesion localization across both modalities, accurate ovarian segmentation is important, as it provides crucial anatomical context~\cite{mittal2025artificial}.  
% In recent years, deep learning methods have substantially advanced computer-aided analysis of female pelvic imaging. However, existing studies remain fragmented: segmentation studies are mostly modality-specific~\cite{lyu2025unsupervised,figueredo2024automatic}, while multimodal efforts mainly focus on disease-level classification~\cite{wang2025human,zhang2025unpaired}. With respect to TVUS images, nnU-Net-based frameworks~\cite{isensee2021nnu} have been adopted for supervised uterus segmentation~\cite{bonevs2024automatic,tank2025automatic}, while domain adaptation techniques have been explored to improve ovarian segmentation across heterogeneous TVUS datasets~\cite{zhao2022mmotu}. In MRI, self-supervised learning has shown promising performance for endometriosis classification~\cite{butler2023effectiveness}. In terms of multimodal learning, Zhang et al. explored the transfer of information from TVUS to MRI to facilitate deep infiltrating endometriosis classification~\cite{zhang2023distilling}. However, multimodal ovary segmentation, particularly cross-modal segmentation, remains largely unexplored. 

Recent studies have advanced Deep Learning for female pelvic image analysis, including modality-specific segmentation in TVUS and MRI~\cite{arjomandi2026prompts,bonevs2024automatic,figueredo2024automatic,lyu2025unsupervised,saleem2025deep,tank2025automatic} and multi-modal learning for endometriosis classification~\cite{butler2023effectiveness,wang2025human,zhang2025unpaired}. 
Nevertheless, ovary segmentation in pelvic MRI remains particularly challenging, as ovaries are small and visually ambiguous among surrounding organs or lesions~\cite{liang2025multi}.
While TVUS often captures more distinctive ovary-specific features, obtaining paired TVUS–MRI examinations for supervised cross-modal learning is difficult in clinical practice~\cite{zhang2023distilling}. Consequently, methods that can effectively transfer anatomical knowledge from TVUS to MRI without requiring paired data are still scarce.

%However, cross-modal ovary segmentation, especially transferring knowledge from TVUS to MRI without paired data, remains underexplored.  
%
% Ovary segmentation in female pelvic MRI is in general a more challenging task than in TVUS, as ovaries are often small and can be easily confused by surrounding organs or lesions~\cite{liang2025multi}.
%Ovary segmentation in pelvic MRI is more challenging than in TVUS because ovaries are small and visually ambiguous among surrounding organs or lesions~\cite{liang2025multi}. 
%Although paired TVUS–MRI data could provide anatomical guidance, such data are difficult to obtain in clinical practice~\cite{zhang2023distilling}. 
% A straightforward approach to address this challenge would be to leverage anatomical information from TVUS to guide MRI segmentation, for example by acquiring paired TVUS–MRI data, performing cross-modal registration, and fusing features across modalities.
% However, in real clinical scenarios, obtaining such paired datasets is highly challenging~\cite{zhang2023distilling}. 

%Therefore, i
%Instead of relying on paired multi-modal fusion, we propose to 
%use unpaired TVUS ovary masks to construct a population-level ovary foreground prior by introducing a dual-branch framework design, which regularizes MRI feature learning through prototype contrastive loss. 
%leverage unpaired TVUS ovary masks to learn a population-level ovarian foreground prior. 
To address this challenge, we propose a cross-modal framework that leverages unpaired TVUS ovary masks to learn a population-level ovarian prior from unpaired TVUS ovary masks and transfers this knowledge to improve MRI ovary segmentation without paired TVUS–MRI data.
We incorporate this prior into a dual-branch framework, where a prototype-based contrastive objective regularizes MRI feature representations toward anatomically meaningful ovarian features.
%
%In general, 
Specifically, our contributions are threefold:
% \begin{enumerate}[label=(\roman*)]
%     \item We propose a dual-branch framework for MRI ovary segmentation built upon MedSAM3~\cite{liu2025medsam3}, a promptable concept segmentation foundation model with strong generalization across anatomical structures and imaging modalities.
%     \item We introduce a population-level ovarian foreground prior constructed from unpaired TVUS ovary masks, providing anatomical guidance for MRI representation learning.
%     \item We propose a prototype-prior contrastive learning strategy that encourages anatomically consistent feature representations across modalities and improves MRI ovary segmentation.
% \end{enumerate}
(i) We propose a dual-branch framework for MRI ovary segmentation built upon MedSAM3~\cite{liu2025medsam3}, a promptable concept segmentation foundation model with strong generalization across anatomical structures and imaging modalities.
(ii) We introduce a population-level ovarian prior constructed from unpaired TVUS ovary masks in form of a prototype bank, providing anatomical guidance for MRI representation learning.
(iii) %We propose a prototype contrastive learning strategy that encourages anatomically consistent feature representations across modalities and improves MRI ovary segmentation.
We propose prototype contrastive learning to enforce cross-modal anatomical consistency and improve MRI ovary segmentation.

%Furthermore, 
%We adapt MedSAM3 to ovary segmentation using Parameter-Efficient Fine-Tuning (PEFT)~\cite{han2024parameter}, enabling learning from unpaired TVUS and MRI datasets.
%To address this gap, we build on MedSAM3~\cite{liu2025medsam3}, a promptable concept segmentation foundation model with strong generalization ability across anatomical structures and imaging modalities. We employ Parameter-Efficient Fine-Tuning (PEFT)~\cite{han2024parameter} to adapt the model to ovary segmentation across unpaired TVUS and MRI data. However, comparing to TVUS, MRI poses additional challenges due to its larger field of view and the ambiguous boundaries between the ovaries and surrounding pelvic structures~\cite{liang2025multi}. To mitigate these difficulties, we introduce a prototype-prior contrastive learning strategy that encourages anatomically consistent feature representations across modalities. 
We systematically evaluate and ablate the proposed framework on public endometriosis-related TVUS and MRI datasets~\cite{liang2025multi,zhao2022mmotu}. 

\section{Methods}
\begin{figure}[t!]
    \centering
    \hspace*{-0.55cm}
    \includegraphics[width=1.05\linewidth]{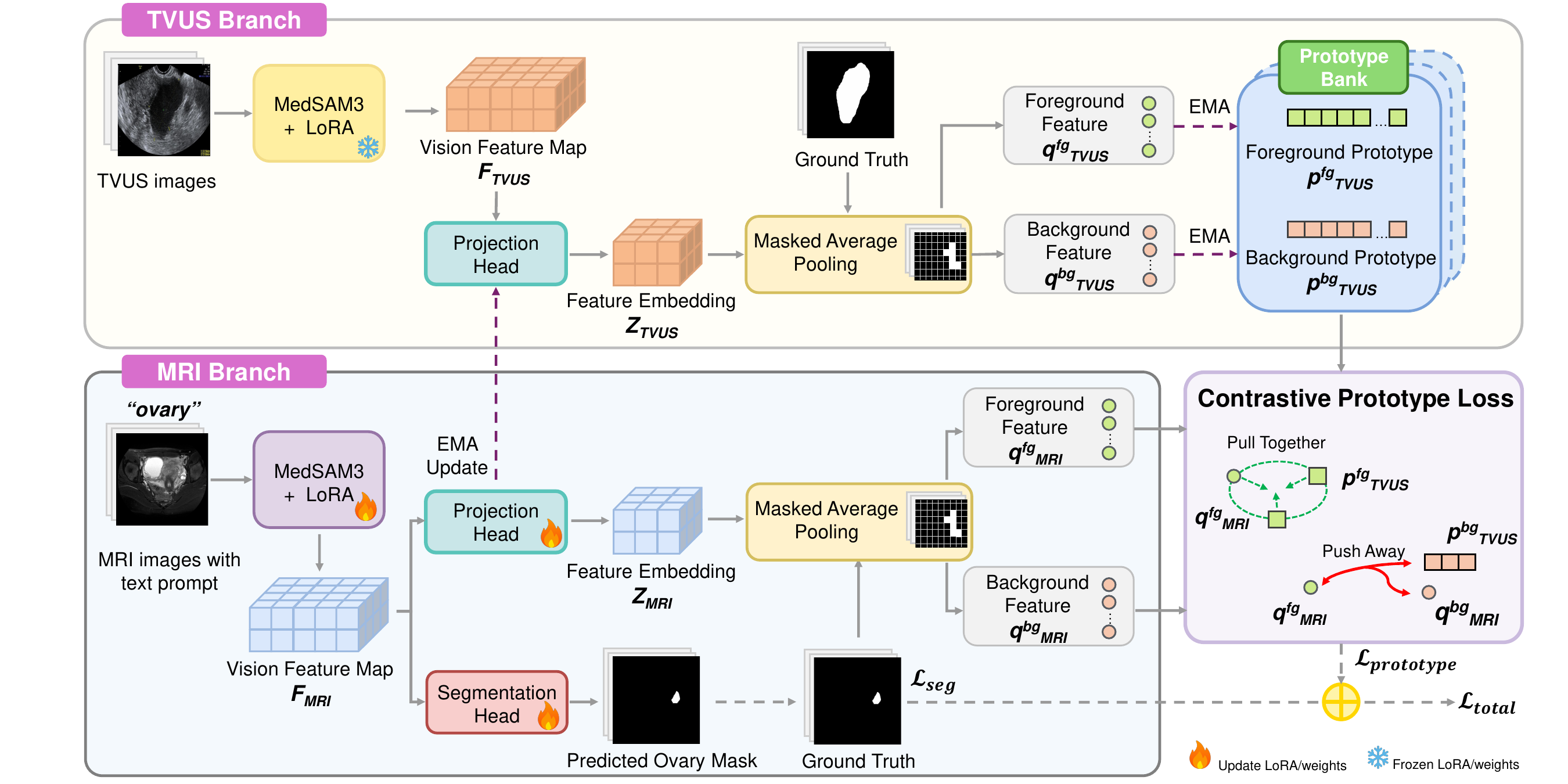}
    \caption{\textbf{Overview of our cross-modal prototype-contrastive learning framework.} Unpaired TVUS and MRI images are encoded in a shared representation space, where prototype contrastive learning promotes cross-modal anatomical consistency. }
    \label{fig:workflow}
\end{figure}
%\subsection{Problem Statement}
%Ovary segmentation in female pelvic MRI is in general a more challenging task than in TVUS, as ovaries are often small and can be easily confused by surrounding organs or lesions. 
%This can also be observed in our preliminary experiments, where both MedSAM3 and the naive SAM~\cite{carion2025sam} show weaker performance on MRI than on TVUS. 
%Given the relatively stronger segmentation performance on TVUS, a straightforward solution would be to utilize TVUS-derived ovarian information to guide MRI segmentation, for instance by collecting patient-level paired TVUS-MRI data, performing cross-modal registration, and applying feature fusion. However, in real clinical scenarios, obtaining such paired datasets is extremely challenging~\cite{zhang2023distilling}. Therefore, instead of relying on paired multimodal fusion, we utilize unpaired TVUS ovary masks to construct a population-level ovary foreground prior by introducing a dual-branch framework design, which regularizes MRI feature learning through prototype contrastive loss. 

To inject anatomical knowledge from TVUS into MRI segmentation, we construct a population-level ovarian foreground prior from unpaired TVUS ovary masks. This prior is incorporated into a dual-branch framework as shown in~\cref{fig:workflow}, where a prototype-based contrastive loss regularizes MRI feature representations towards anatomically consistent ovarian features. 
% The overall framework for the proposed method is shown in~\cref{fig:workflow}.
\Cref{sec:backbone} describes the proposed dual-branch architecture and its MedSAM3~\cite{liu2025medsam3} backbone, while \cref{sec:prior} presents the construction of the population-level ovarian prior from unpaired TVUS ovary masks. 
Finally, \cref{sec:contrastive} introduces the proposed prototype-prior contrastive learning strategy. 

\subsection{Dual-branch Framework and MedSAM3 Backbone}
\label{sec:backbone}
%In the proposed framework, dual branch. For each branch we use MedSAM3 is adopted to be the backbone model. 
In the proposed dual-branch framework, MedSAM3~\cite{liu2025medsam3} serves as the backbone model for both branches. The MRI and TVUS branches process their respective inputs independently while sharing a common architecture, enabling the learning of modality-specific features and cross-modal anatomical representations.

We adapt MedSAM3 to ovary segmentation using Parameter-Efficient Fine-Tuning (PEFT)~\cite{han2024parameter}. %, enabling learning from unpaired TVUS and MRI datasets.
Building on the pretrained SAM3~\cite{carion2025sam}, MedSAM3 utilized Low-rank Adaptation (LoRA) to finetune the original model on a set of medical terminologies and images. Therefore, for a given input image, MedSAM3's Vision Transformer (ViT)-based vision backbone produces LoRA-adapted visual representations that better capture medical image semantics. Specifically, the vision backbone generates a multi-scale feature representation through a Feature Pyramid Network (FPN). 

In our framework, for each modality branch and input batch of size $B$, we use the final FPN output as the visual feature representation $F\in\mathbb{R}^{B\times C\times H_{f}\times W_{f}}$
%, since it contains the high-level semantically enriched visual information for subsequent prototype alignment. 
, as it captures high-level semantic information that facilitates subsequent prototype alignment.
In the TVUS branch, the backbone and LoRA modules are frozen after warm up, while the projection head is updated via an \ac{ema} from the MRI branch.
In the MRI branch, the LoRA modules and projection head are trainable. During inference, the projection heads are removed, and segmentation is performed using MedSAM3 with the learned MRI LoRA modules.

\subsection{Prototype Bank}
\label{sec:prior}
We further employ a lightweight two-layer $1\times1$ convolutional projection head with GroupNorm, ReLU, and channel-wise L2 normalization to map the backbone features $F$ into a normalized embedding $Z\in\mathbb{R}^{B\times C_{\mathrm{proj}}\times H_f\times W_f}$. The ovary mask is then resized to the corresponding feature resolution and used to extract the foreground features $q^{fg}\in \mathbb{R}^{B\times C_{proj}}$ and background features $q^{bg}\in \mathbb{R}^{B\times C_{proj}}$ from $Z$ via masked average pooling. Hence, each input image is represented by a pooled foreground feature vector and a pooled background feature vector. 
%To create a prototype prior $p$ and therefore reduce the batch-wise noise and stabilize the ovary prior, we utilized a prototype memory bank to the TVUS branch and update the bank with Exponential Moving Average (EMA) with the extracted pooled foreground and background feature vectors from the TVUS inputs. 
To reduce batch-wise noise in the ovary representation, we propose to create a prototype memory bank using a stable prototype prior $p$ in the TVUS branch.
The prototype memory bank stores $B$ foreground prototypes $p_{\mathrm{TVUS}}^{fg}$ and $B$ background prototypes $p_{\mathrm{TVUS}}^{bg}$, where $B$ corresponds to the batch size. 
During training, the prototypes are updated using an Exponential Moving Average (EMA) of the foreground and background features extracted from the TVUS inputs.
Each prototype is updated according to
\begin{equation}
    p_{\mathrm{TVUS}}^{fg}(t) = \alpha \, p_{\mathrm{TVUS}}^{fg}(t-1) + (1-\alpha) \, q_{\mathrm{TVUS}}^{fg}(t) \; ,
\end{equation} 

%\begin{equation}
%    p_{\mathrm{TVUS}}^{fg}(t) = \alpha \, p_{\mathrm{TVUS}}^{fg}(t-1) + (1-\alpha) \, q_{\mathrm{TVUS}}^{fg}(t) \; ,
%\end{equation} 

\noindent
with an analogous update applied to the background prototypes.
% \begin{equation}
%     q_{\mathrm{TVUS},t}^{fg} = \alpha q_{\mathrm{TVUS},t-1}^{fg}+(1-\alpha)q_{\mathrm{TVUS},t}^{fg} \; ,
% \end{equation} 
% while the background prototypes: 
% \begin{equation}
%     q_{\mathrm{TVUS},t}^{bg} = \alpha q_{\mathrm{TVUS},t-1}^{bg}+(1-\alpha)q_{\mathrm{TVUS},t}^{bg} \; ,
% \end{equation} 
Here, $\alpha$ denotes the momentum coefficient and $t$ the training step. 
The prototype bank is initialized as $p_{\mathrm{TVUS}}^{fg}(0) = q_{\mathrm{TVUS}}^{fg}(0)$ at the first step.

\subsection{Prototype Contrastive Learning}
\label{sec:contrastive}
In the proposed framework, the TVUS foreground and background prototypes are introduced to regularize the MRI branch's feature learning. Since the TVUS and MRI datasets are unpaired, we do not enforce spatial correspondence between the two modalities. Instead, based on the shared anatomical semantics in both modalities, we encourage the MRI ovary foreground features to move closer to the TVUS ovary foreground prototypes, while being pushed away from the TVUS background prototypes and MRI background features. This is achieved by using the InfoNCE contrastive loss~\cite{oord2018representation} defined as: 
{\small
\begin{equation}
    \label{eq:infonce}
    \begin{aligned}
    \mathcal{L}_{proto}&=\\
    -\log&
    \frac{
        \exp\left(\mathrm{sim}\left(q_{\mathrm{MRI}}^{fg}, p_{\mathrm{TVUS}}^{fg}\right)/\tau\right)
    }
    {
        \exp\left(\mathrm{sim}\left(q_{\mathrm{MRI}}^{fg}, p_{\mathrm{TVUS}}^{fg}\right)/\tau\right)
        +
        \sum_{z \in \left\{p_{\mathrm{TVUS}}^{bg}, q_{\mathrm{MRI}}^{bg}\right\}}
        \exp\left(\mathrm{sim}\left(q_{\mathrm{MRI}}^{fg}, z\right)/\tau\right)
    } \; ,
    \end{aligned}
\end{equation}
}
\leavevmode\par\noindent
where $sim(\cdot,\cdot)$ denotes cosine similarity of two embeddings and $\tau$ is the temperature parameter, $p_{\mathrm{TVUS}}^{fg}$ is the positive anchor and $z \in \{p_{\mathrm{TVUS}}^{bg}, q_{\mathrm{MRI}}^{bg}\}$ is the negative anchor. 
%The contrastive loss is weighted with a parameter $\lambda$ and then combined with the original segmentation loss $\mathcal{L}_{seg}$ to form the total training objective as defined in ~\cref{eq:loss}:  
The contrastive loss is weighted by $\lambda$ and combined with the segmentation loss $\mathcal{L}_{seg}$ to form the total loss:
\begin{equation}
\label{eq:loss}
    \mathcal{L}_{\mathrm{total}} = \mathcal{L}_{seg} + \lambda \, \mathcal{L}_{proto} \; . 
\end{equation}

\section{Experiments and Results}
% We evaluate the proposed framework for ovary segmentation in female pelvic MRI. 
% We first describe the datasets (Section~\ref{sec:data}) and implementation details (Section~\ref{sec:implementation}), followed by comparisons with state-of-the-art methods in Section~\ref{sec:baselines}. 
% Quantitative and qualitative results to assess segmentation accuracy and robustness are presented in Section~\ref{sec:results}. 
% Finally, we conduct an ablation studies to analyze the contribution of each proposed component (Section~\ref{sec:additional_exp}).
% This section evaluates the proposed framework for ovary segmentation in female pelvic MRI. We begin by introducing the datasets and implementation details, as well as the state-of-the-art methods used for comparison. Building on this setup, we then present the quantitative and qualitative results to assess the overall performance of the proposed framework. Finally, we conduct ablation studies to further investigate the contribution of each component. 
We evaluate the proposed framework for ovary segmentation in female pelvic MRI. 
~\cref{sec:data} describes the datasets, ~\cref{sec:implementation} the implementation details, and~\cref{sec:baselines} the comparison with state-of-the-art methods. Quantitative and qualitative results are presented in~\cref{sec:results}, followed by ablation studies in~\cref{sec:additional_exp}.

\subsection{Datasets}
\label{sec:data}
To train and evaluate the proposed method, we used two publicly accessible female pelvic imaging datasets. For the TVUS branch, we used the OTU-2D dataset~\cite{zhao2022mmotu}, which originally contains ultrasound images from 247 patients from eight different categories with corresponding ovary segmentation masks. To align with the endometriosis task, we selected images labeled as endometriomas (chocolate cysts) and normal ovaries. For the MRI branch, we used the UT-EndoMRI~\cite{liang2025multi} dataset, consisting of multi-contrast MRI images and ovary segmentation masks from 81 female patients with endometriosis. 
We used the T2-weighted fat-satur\-ation (T2FS) volumes with the corresponding ovarian annotations from one obstetrician-gynecologist assistant supervised by an experienced gynecologist. Of note, only a single ovary label was provided per volume.
%In terms of preprocessing, the 3D 
Volumes and masks were resampled to 1.0$\times$1.0$\times$1.0~mm, and sliced into 2D axial frames. 
%Every frame containing an ovary mask is then selected from the volume and resized to 1024$\times$1024. Furthermore, we apply minmax-normalization to the pixel intensities. 
2D slices containing ovary annotations were selected. Both MRI and ultrasound images were resampled and bilinear interpolated to 1008$\times$1008~px, and min-max normalized prior to training and inference of SAM3~\cite{carion2025sam}.
%
% Both datasets were split into training/validation/test sets at a ration of 60\%/10\%/30\%. 
% Importantly, we ensure there was no patient overlap between training/validation sets and test set for both dataset. In total, the dataset consists of 669 TVUS images and 285 MRI images. 
%The TVUS and MRI datasets were independently divided into training, validation, and test sets using a 60\%/10\%/30\% split. 
%Overall, the datasets contain 669 TVUS images and 285 MRI 2D slices. 
%To prevent data leakage and ensure unbiased evaluation, all splits were performed at the patient level. 

Overall, the TVUS and MRI datasets comprise 669 TVUS images and 285 MRI slices, respectively. 
Both datasets were independently divided into training, validation, and test sets with a 60\%/10\%/30\% split. 
To prevent data leakage and ensure unbiased evaluation, all splits were performed at the patient level.
%, with no overlap between the training/validation and test cohorts for either dataset.
% In terms of postprocessing, we adopted the default SAM3 setting with non-maximum-suppression (NMS) and set the IoU threshold to 0.5. 

\subsection{Implementation Details}
\label{sec:implementation}
% During training phase, we first train the TVUS model branch with 3 warm-up epochs to stabilize the feature space for efficient prototype extraction. This is due to the fact that the original MedSAM3 is not being trained on any ovary-related data~\cite{liu2025medsam3}, hence its naive performance on segmenting ovary on TVUS is not sufficient. 
%Since the original MedSAM3 is not being specifically trained or finetuned on ovary-related data and the segmentation performance on TVUS ovary data is not sufficient, we first train the TVUS model branch with 3 warm-up epochs to stabilize the feature space for efficient prototype extraction. 
% Since MedSAM3 was not originally finetuned on ovary-specific data and exhibited limited segmentation performance on TVUS ovary images, we first pre-trained the TVUS branch's LoRA weights independently for three warm-up epochs. This initialization stage stabilizes the feature space and enables the extraction of more reliable prototypes for subsequent cross-modal learning.
% The TVUS model was then frozen and the MRI branch is trained on MRI data and with TVUS-derived prototypes for 10 epochs, including the corresponding projection layers. 
Since MedSAM3 was not originally fine-tuned on ovary-specific data and exhibited limited segmentation performance on TVUS ovary images, we first pre-trained the TVUS branch’s LoRA weights independently for three warm-up epochs. This initialization stage stabilized the feature space and enabled the extraction of more reliable prototypes for subsequent cross-modal learning. 
The TVUS model was then frozen, and the MRI branch, including its projection layers, was trained for 10 epochs on MRI data using TVUS-derived prototypes. 
% During finetuning, bounding boxes were derived from the ground-truth masks to compute the bounding box components of the segmentation loss. 
% In both aforementioned phases, we used batch size $B$ as 8 images per modality (learning rate: 1e-5, optimizer: AdamW). 
% In addition, we set the $\lambda$ to be 0.015 to weight the prototype loss, and the $\alpha$ in EMA as 0.95. In terms of finetuning, the LoRA rank is set to be 16 and LoRA is applied to every SAM3 backbone. 
For both training phases we used AdamW (lr=$10^{-5}$), a batch size of 8 per modality, $\lambda=0.015$, and \ac{ema} momentum $\alpha=0.95$. LoRA (rank 16) was applied to all backbones.

For evaluation, we used the mean Dice coefficient (mDICE) over all images for segmentation and mean Average Precision (mAP, IoU thresholds from 0.50 to 0.95) over all images for organ bounding box detection as metrics. To enable MedSAM3 and SAM3's text-prompt-based segmentation capability~\cite{liu2025medsam3,carion2025sam}, we used a fixed text prompt ("ovary") for all image samples.
During inference, the default SAM3 non-maximum suppression (NMS) and IoU threshold settings were used~\cite{carion2025sam}. 
To address the issue that only a single (one-sided) ovary label was provided per volume, loss computation and quantitative evaluation were restricted to the side for which a reference annotation was available. This allowed to feed the entire image to the model to provide global anatomical context but avoided penalizing models that correctly predicted bilateral ovary structures. 

All training and evaluation runs were executed on 2 NVIDIA A100 40GB GPUs with Python 3.12 and PyTorch 2.11.

\subsection{Baselines}
\label{sec:baselines}
%We compare the proposed prototype-regularized framework with several SAM-based and non-SAM baselines. 
% We compare the proposed prototype-regularized framework against foundation model (FM)-based baselines and a representative non-FM baseline.
We compare the proposed prototype-regularized framework with several baseline methods, including foundation models and conventional supervised approaches.
First, we evaluate the naive SAM3~\cite{carion2025sam} and MedSAM3~\cite{liu2025medsam3} models in a zero-shot setting% and the models are directly applied for inference on the MRI test set. 
, where the pretrained models are directly applied to the MRI test set without further training.
Second, we conduct full LoRA fine-tuning (FT) for both SAM3 and MedSAM3 using only MRI segmentation supervision. 
%Third, we apply the proposed prototype contrastive loss to both SAM3 and MedSAM3 to investigate whether the unpaired TVUS-derived foreground prior can further regularize MRI ovary segmentation. 
In addition, we trained an nnU-Net~\cite{isensee2021nnu} with its 2D configuration on the same MRI dataset as a fully supervised segmentation baseline.  

\subsection{Quantitative and Qualitative Results}
\label{sec:results}
\begin{figure}[t!]
    \centering
    \includegraphics[width=0.95\linewidth]{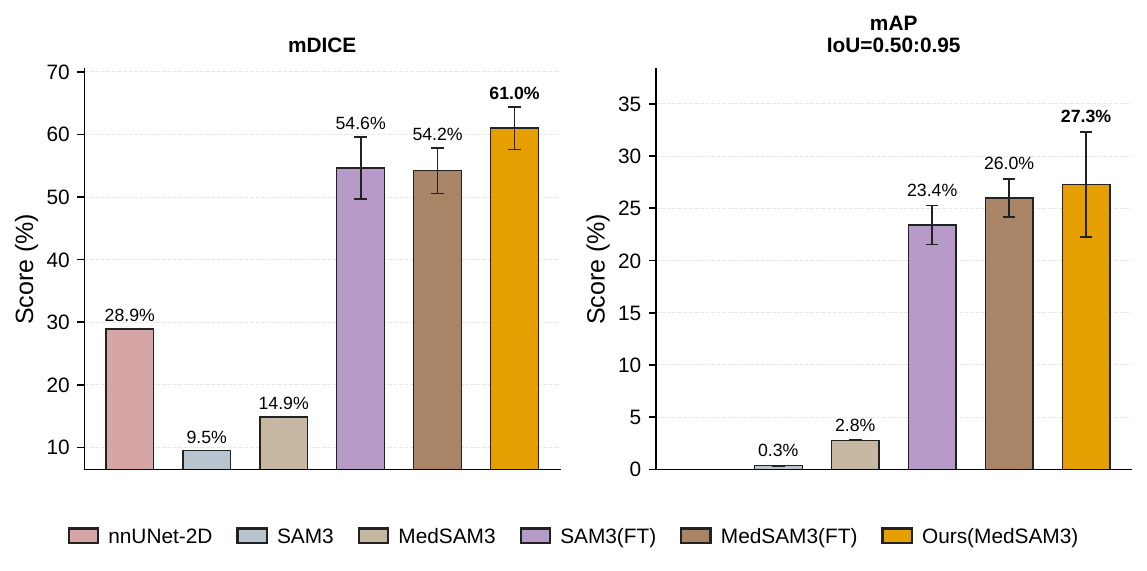}
    \vspace{-0.3cm}
    \caption{\textbf{Performance comparison between proposed methods and baselines on MRI ovary segmentation.} The best performance is bold. ``FT'' refers to the fully supervised LoRA finetuning on the MRI dataset. ``Ours'' denotes the proposed dual-branch approach with a MedSAM3 backbone.} %model trained with the proposed prototype contrastive loss. }
    \label{fig:quantitative}
\end{figure}
\noindent\Cref{fig:quantitative} presents the quantitative comparison between the proposed cross-modality prototype-contrastive learning framework and baseline models for MRI ovary segmentation. The zero-shot SAM3 and MedSAM3 models show limited generalization ability on the MRI test set, indicating that direct transfer from natural-image or medical-image foundation models remains insufficient for this challenging segmentation task. The 2D nnU-Net baseline achieves an mDICE of 28.9\%, but its performance is still clearly lower than that of the SAM3-based fine-tuning approaches. Full LoRA fine-tuning substantially improves the segmentation performance of both SAM3 and MedSAM3 models. 
%Notably, incorporating the proposed TVUS-guided prototype prior brings additional improvements for MedSAM3.
Notably, our proposed dual-branch approach consistently outperforms the LORA-finetuned MedSAM3 baseline, demonstrating the benefit of incorporating TVUS-based prototype priors.
Specifically, compared with full fine-tuning, MedSAM3 equipped with the prototype prior obtains the best segmentation performance and object detection ability, reaching the highest mDICE of 61.0\% and the best mAP score of 27.3\%.  

\begin{figure}[t!]
\centering
\setlength{\tabcolsep}{2pt}
\renewcommand{\arraystretch}{0.85}

\resizebox{\linewidth}{!}{
\begin{tabular}{@{} C{1.2cm} *{7}{C{2.0cm}} @{}}
 & 
\textbf{GT} &
\textbf{nnU-Net} &
\textbf{SAM3} &
\textbf{MedSAM3} &
\shortstack{\textbf{SAM3}\\\textbf{(FT)}} &
\shortstack{\textbf{MedSAM3}\\\textbf{(FT)}} &
\textbf{Ours} \\

\textbf{D2-009} &
\includegraphics[width=2.05cm]{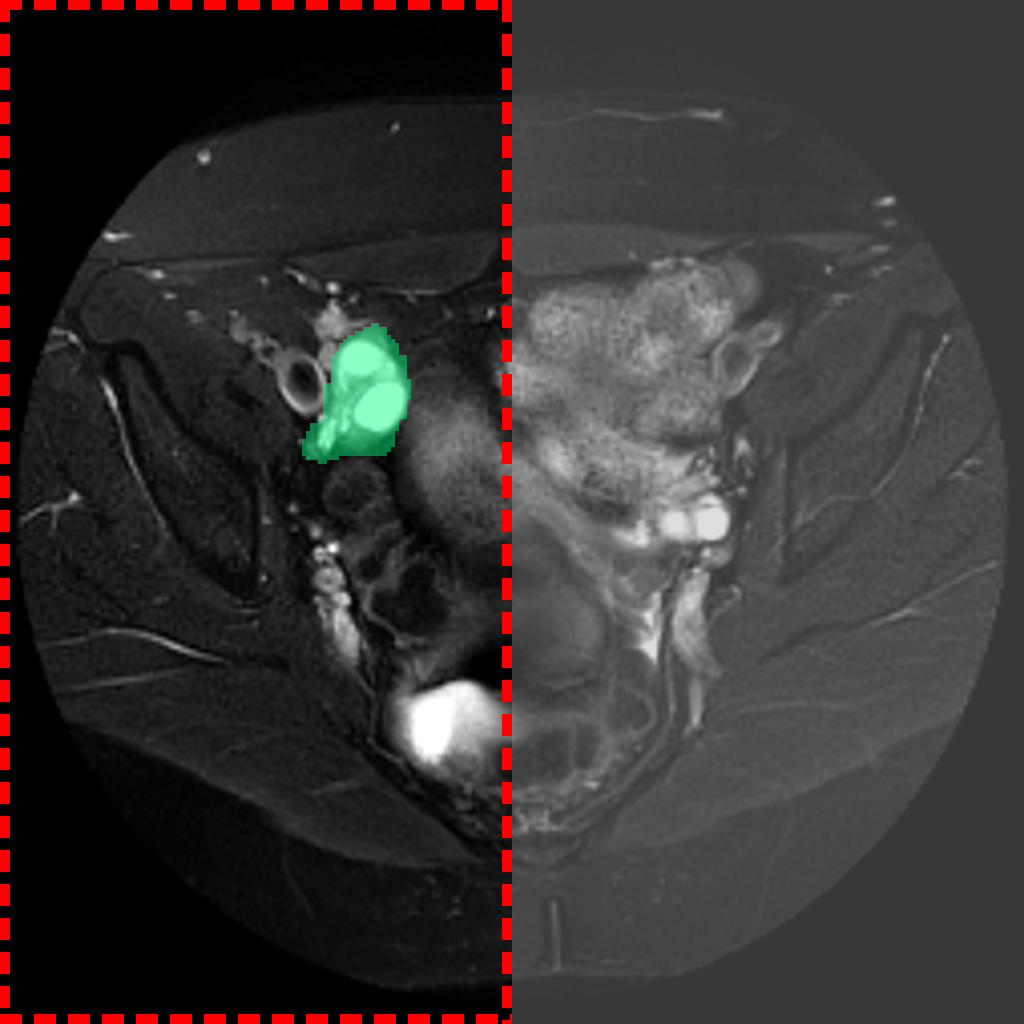} &
\includegraphics[width=2.05cm]{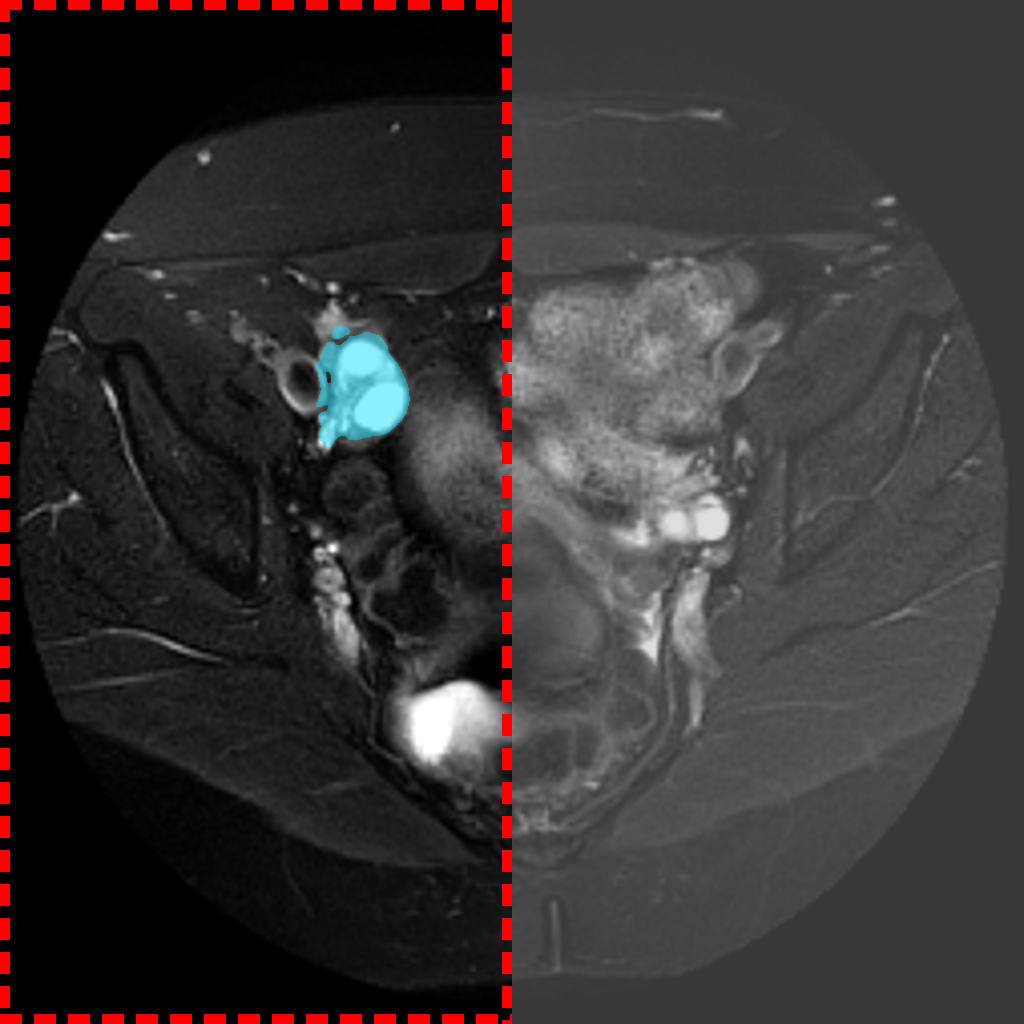} &
\includegraphics[width=2.05cm]{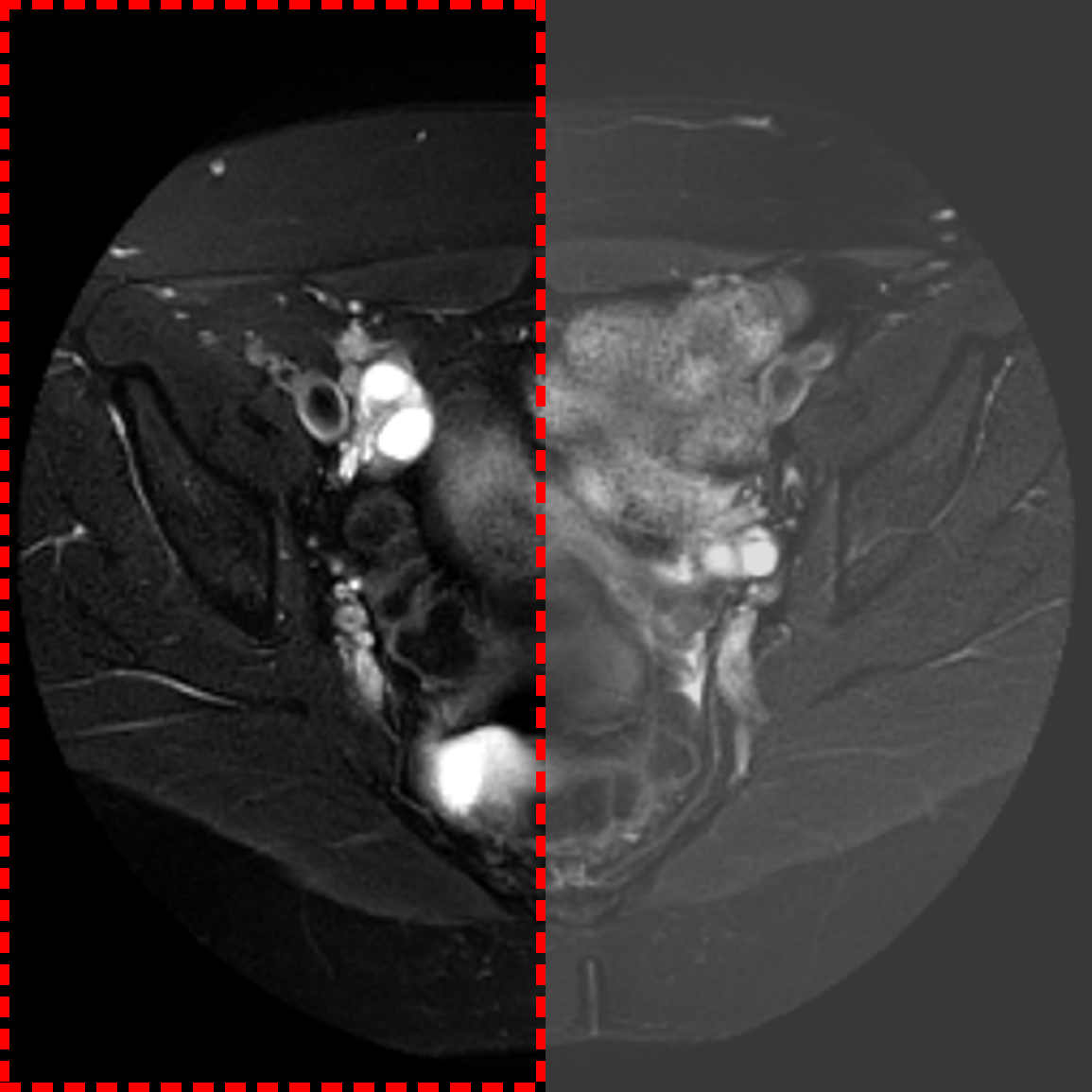} &
\includegraphics[width=2.05cm]{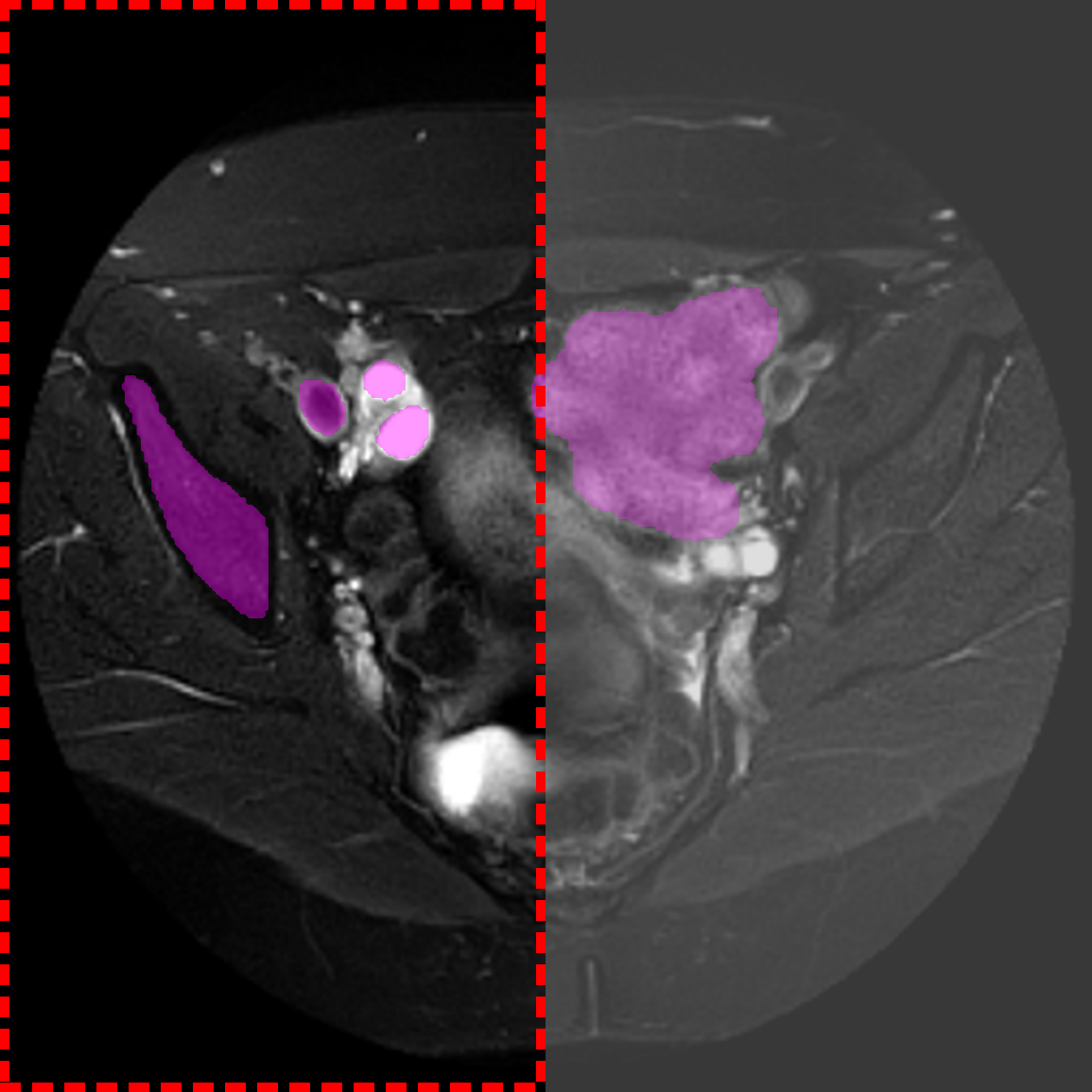} &
\includegraphics[width=2.05cm]{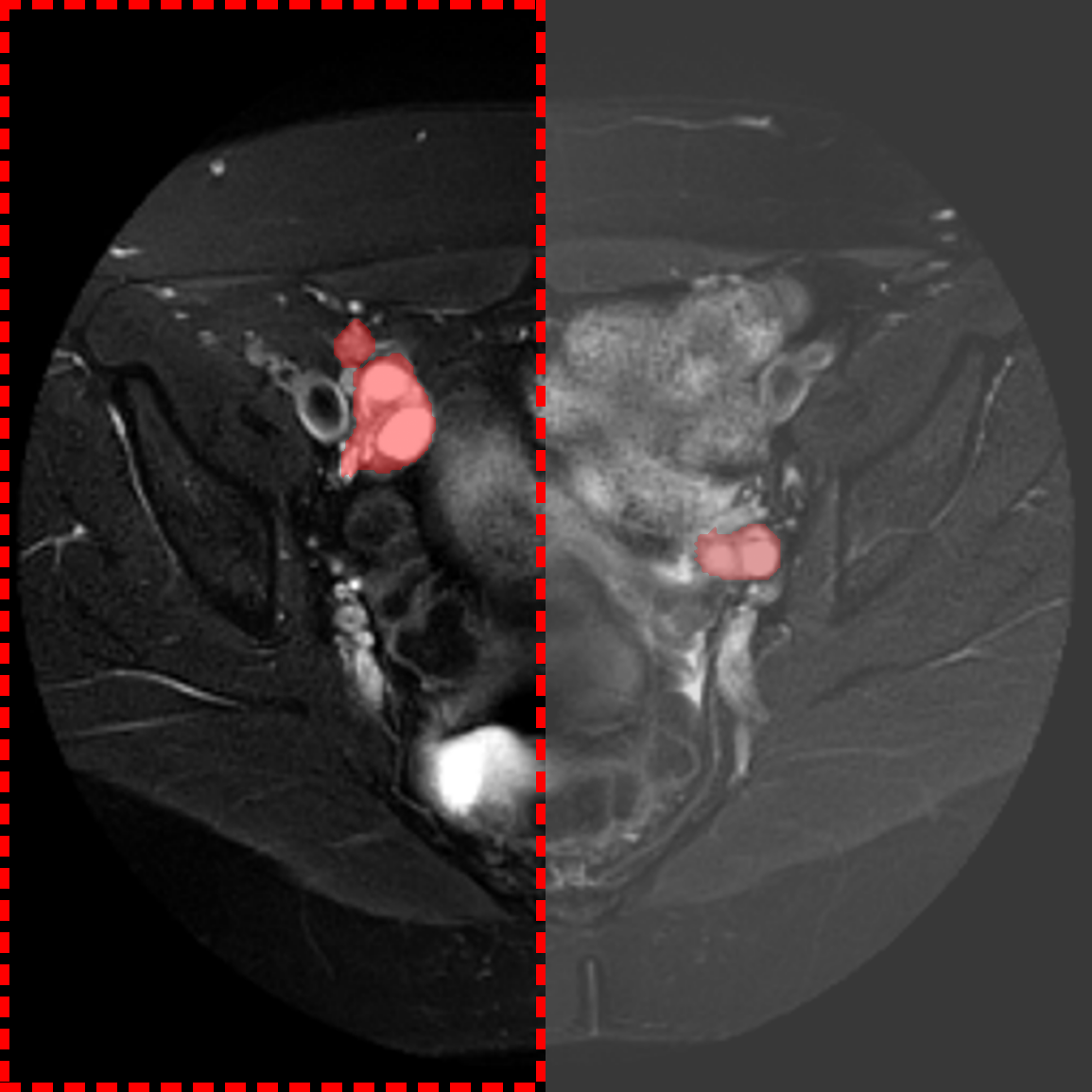} &
\includegraphics[width=2.05cm]{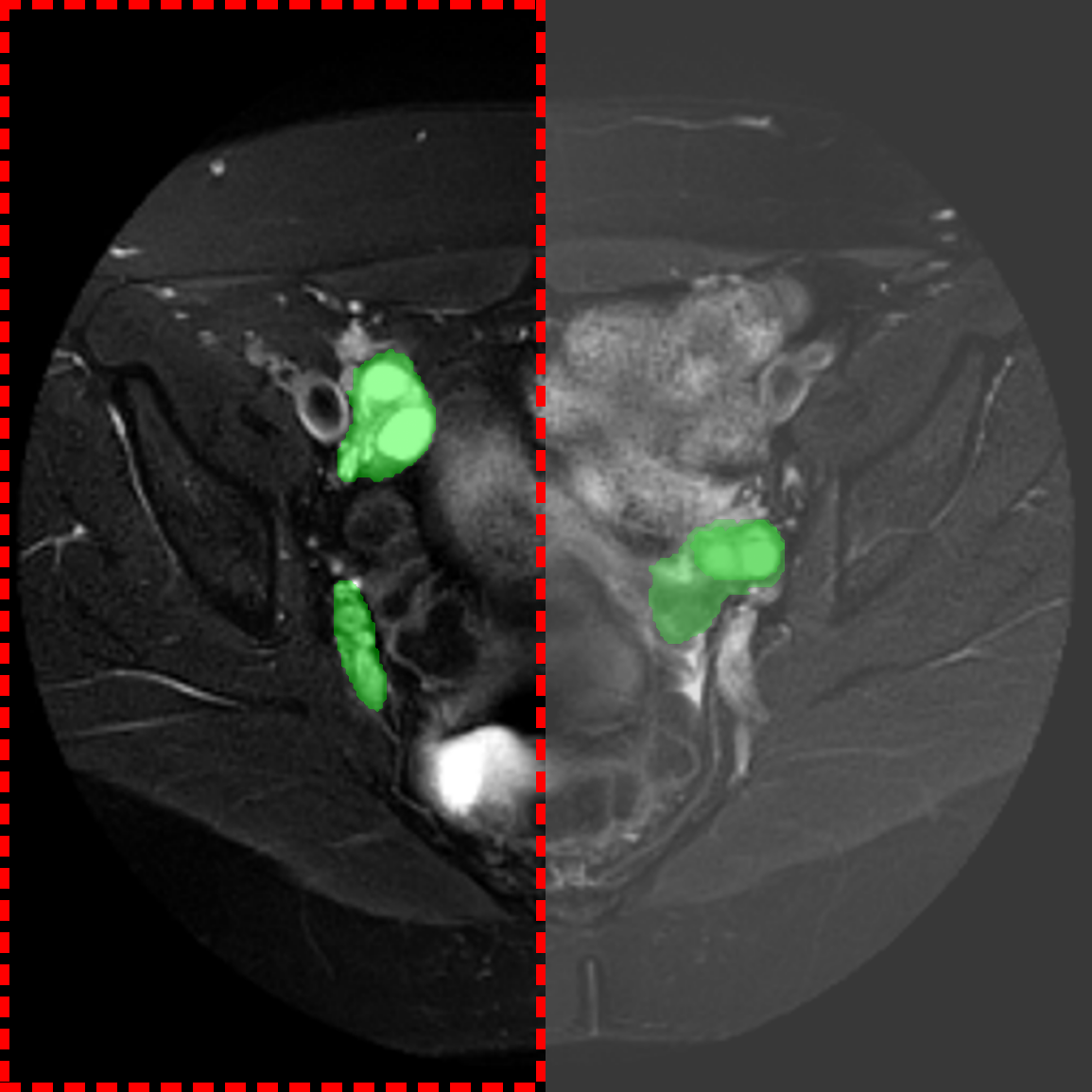} &
\includegraphics[width=2.05cm]{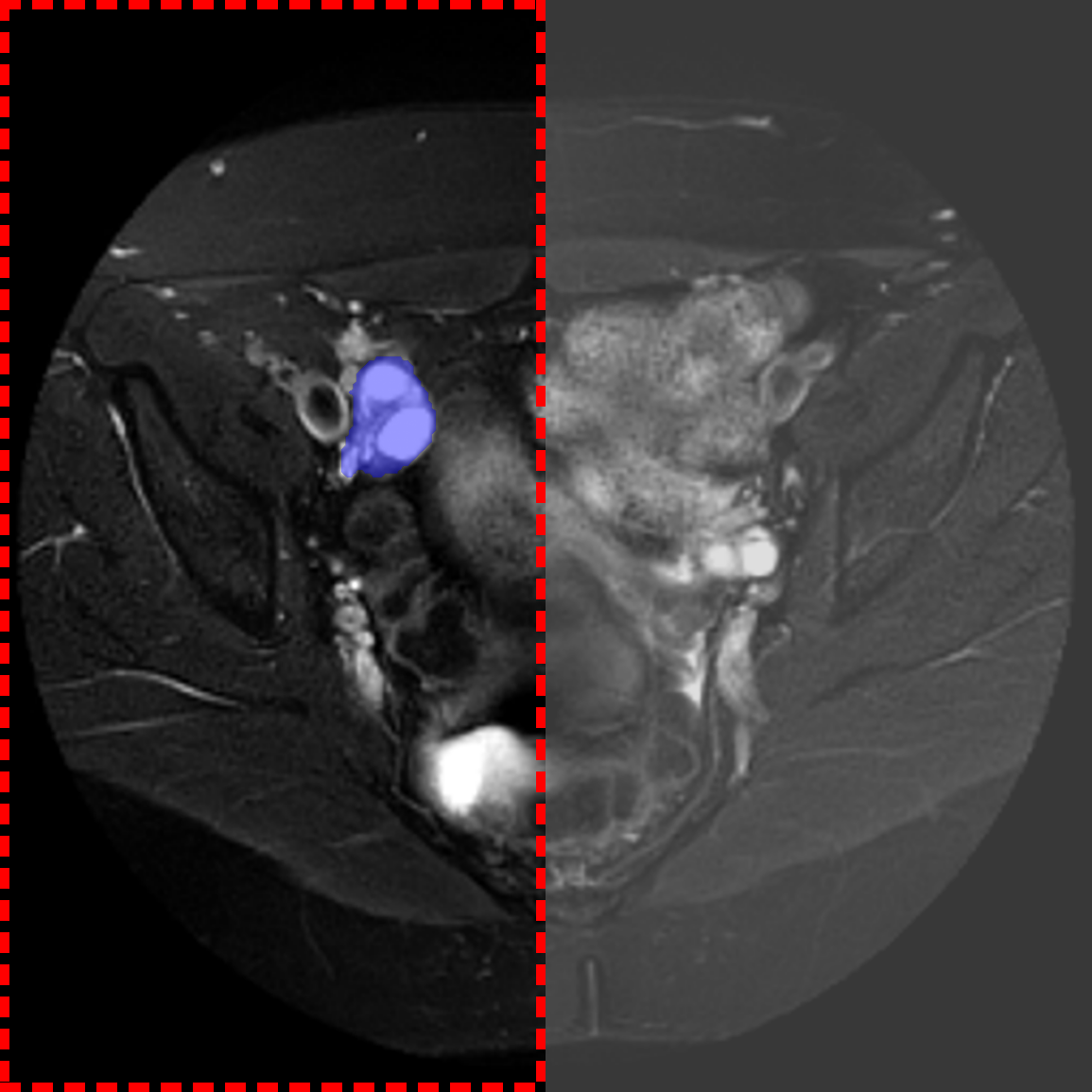} \\

\textbf{D2-054} &
\includegraphics[width=2.05cm]{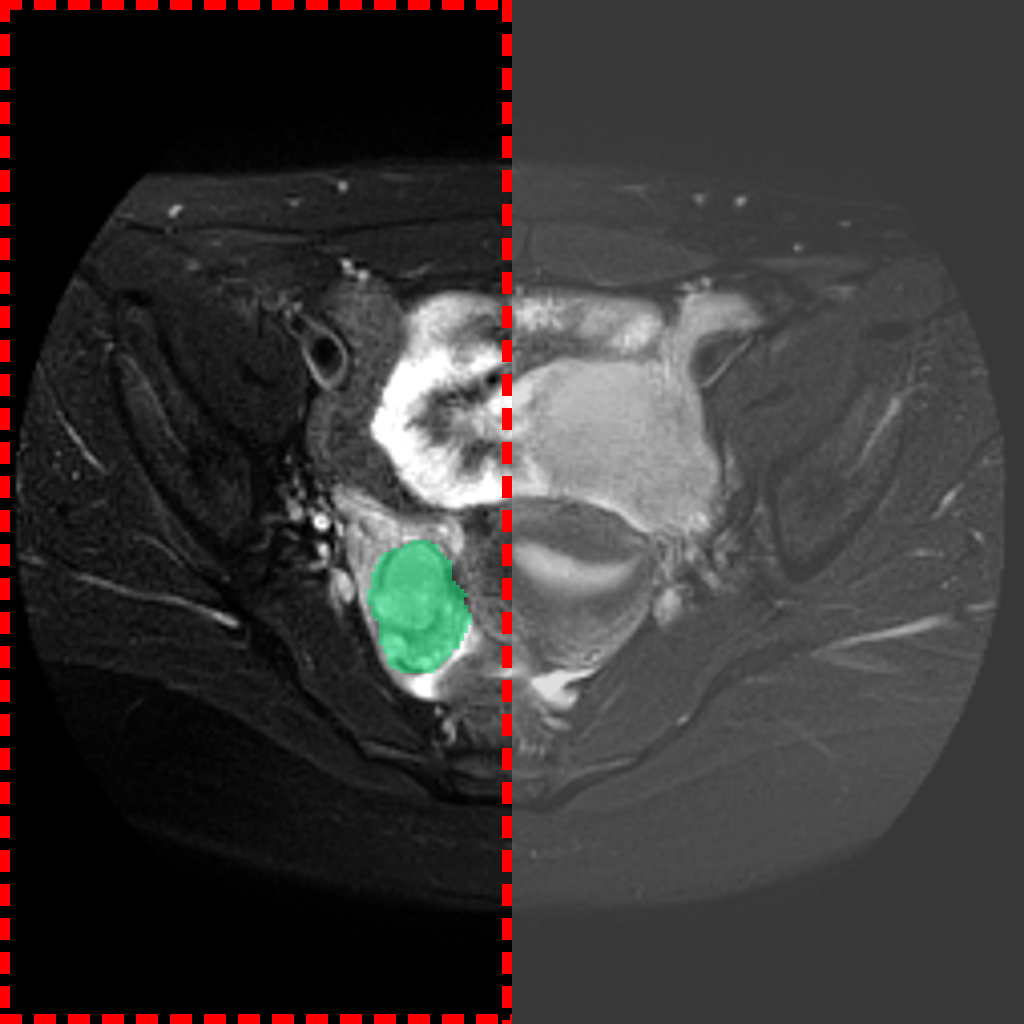} &
\includegraphics[width=2.05cm]{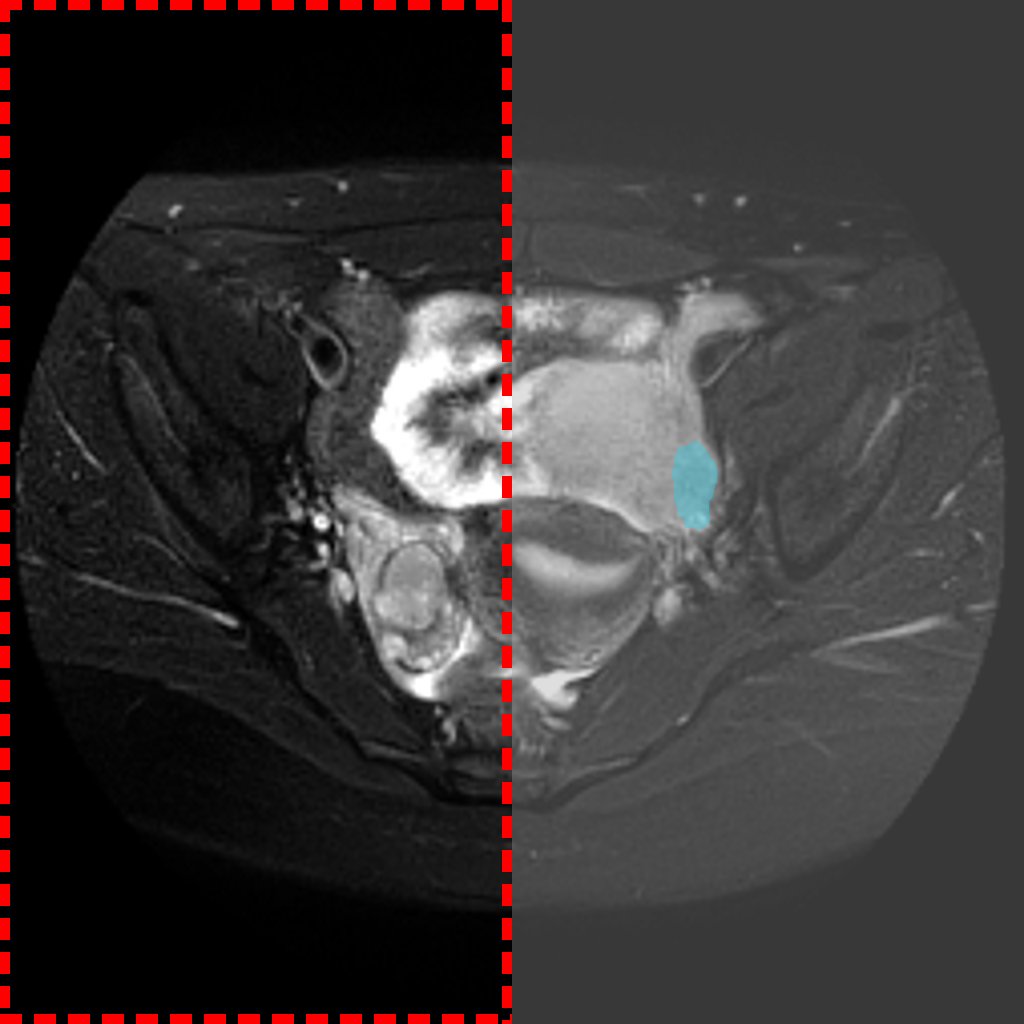} &
\includegraphics[width=2.05cm]{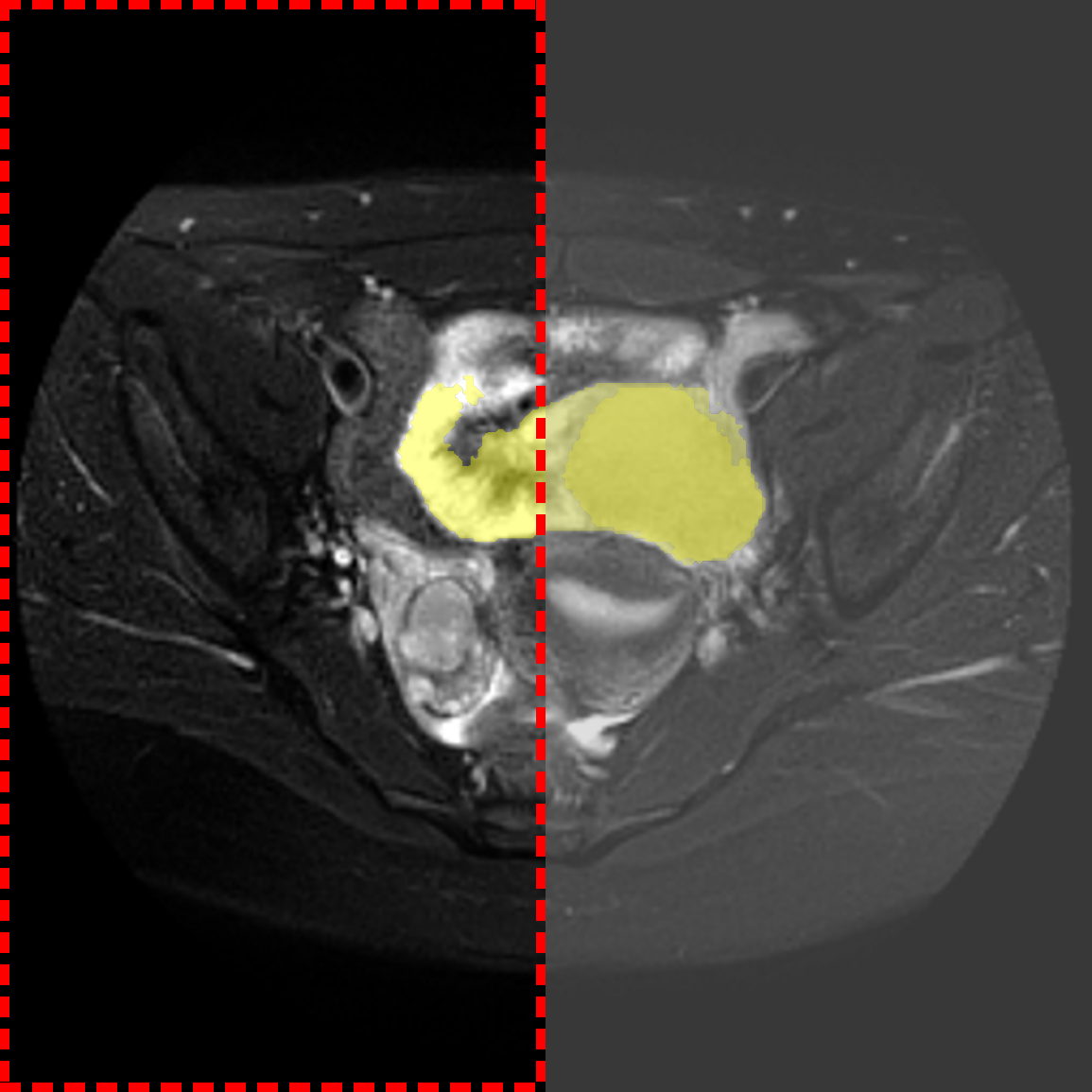} &
\includegraphics[width=2.05cm]{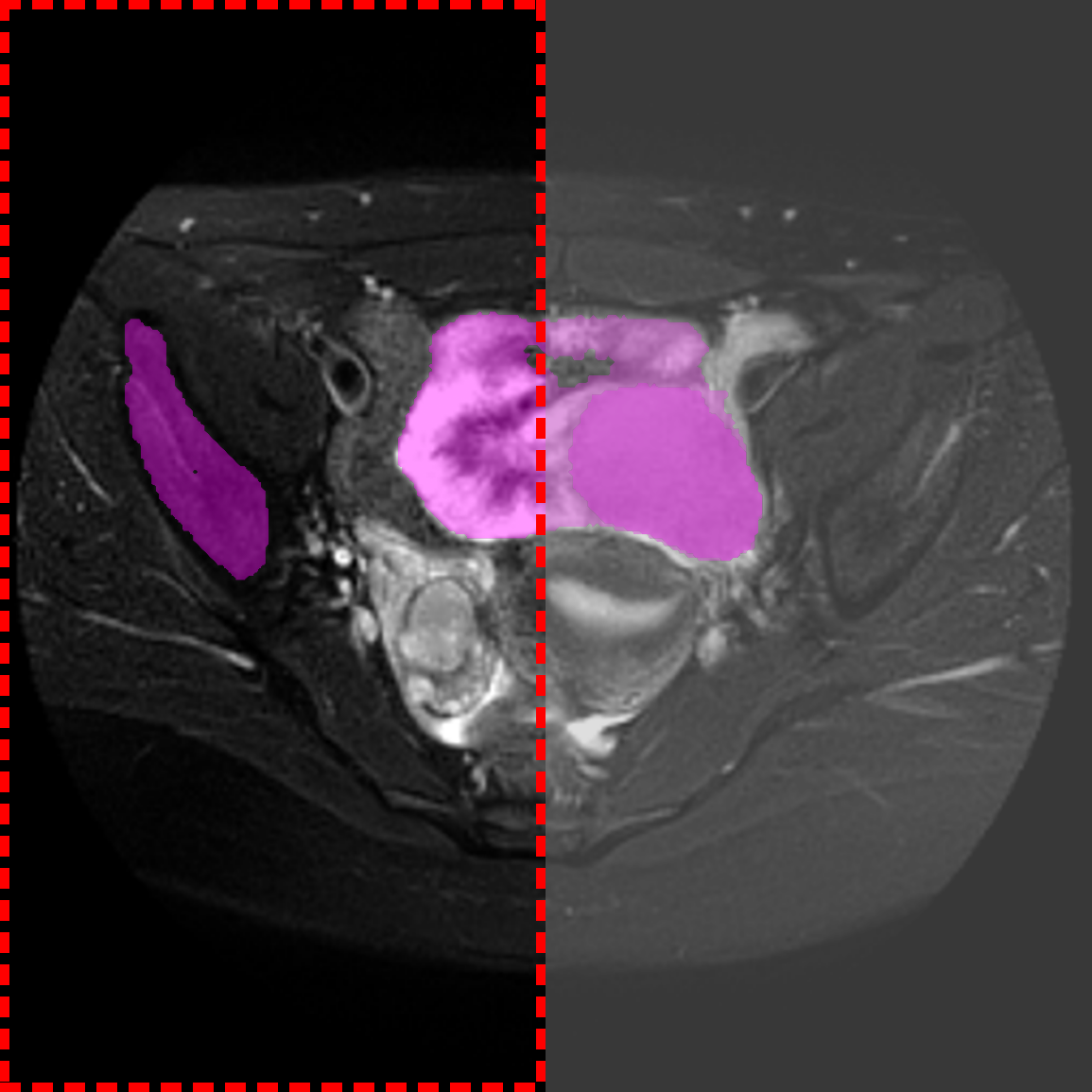} &
\includegraphics[width=2.05cm]{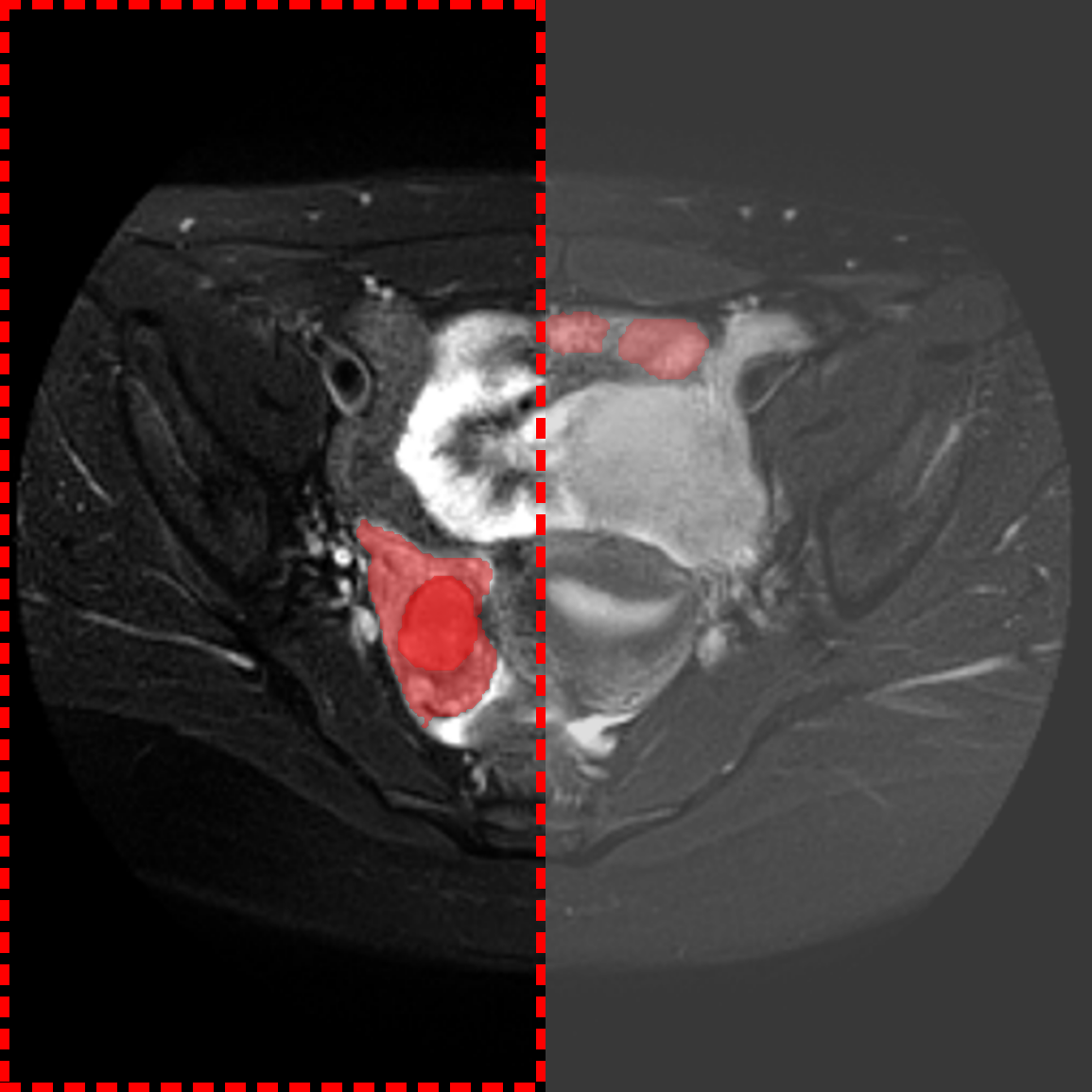} &
\includegraphics[width=2.05cm]{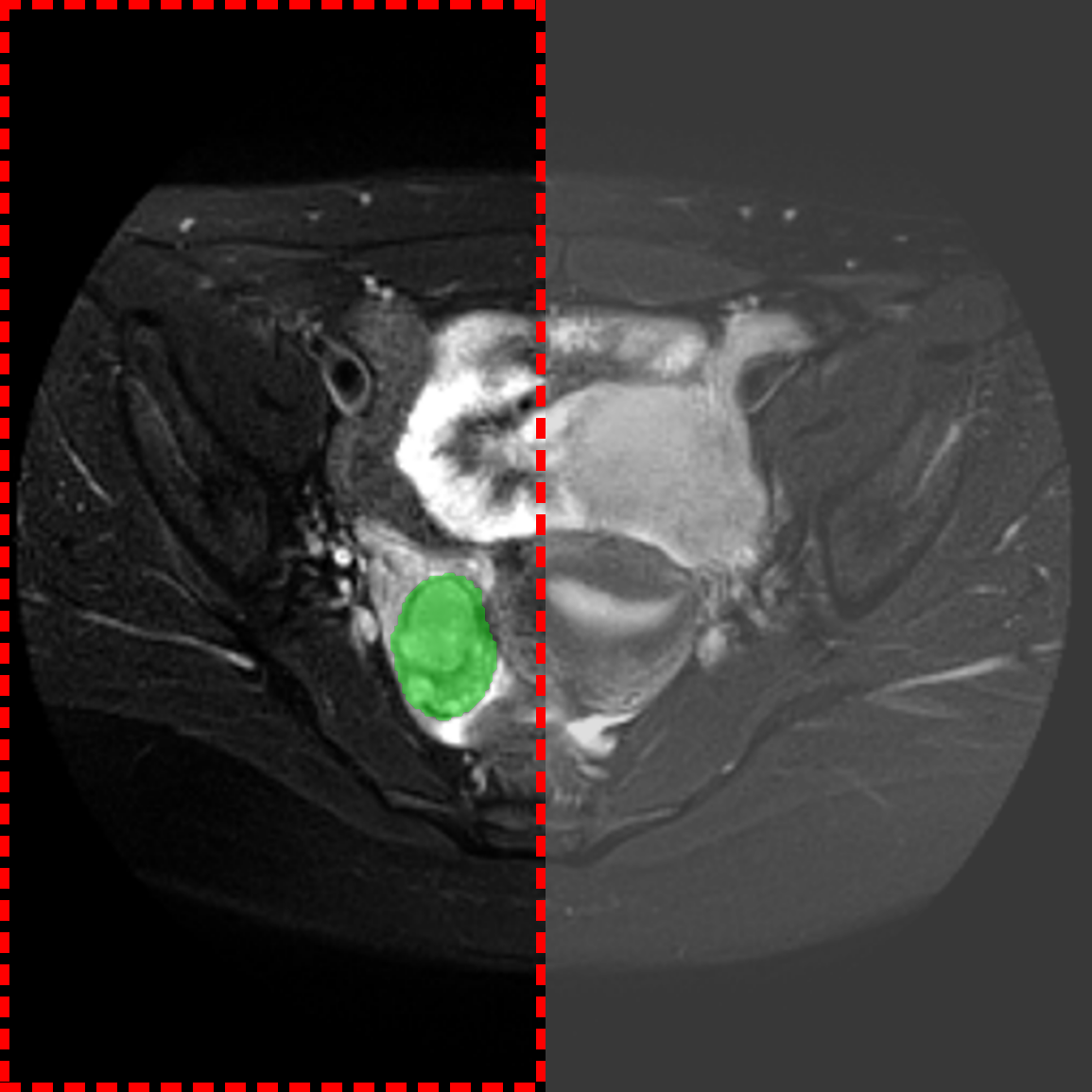} &
\includegraphics[width=2.05cm]{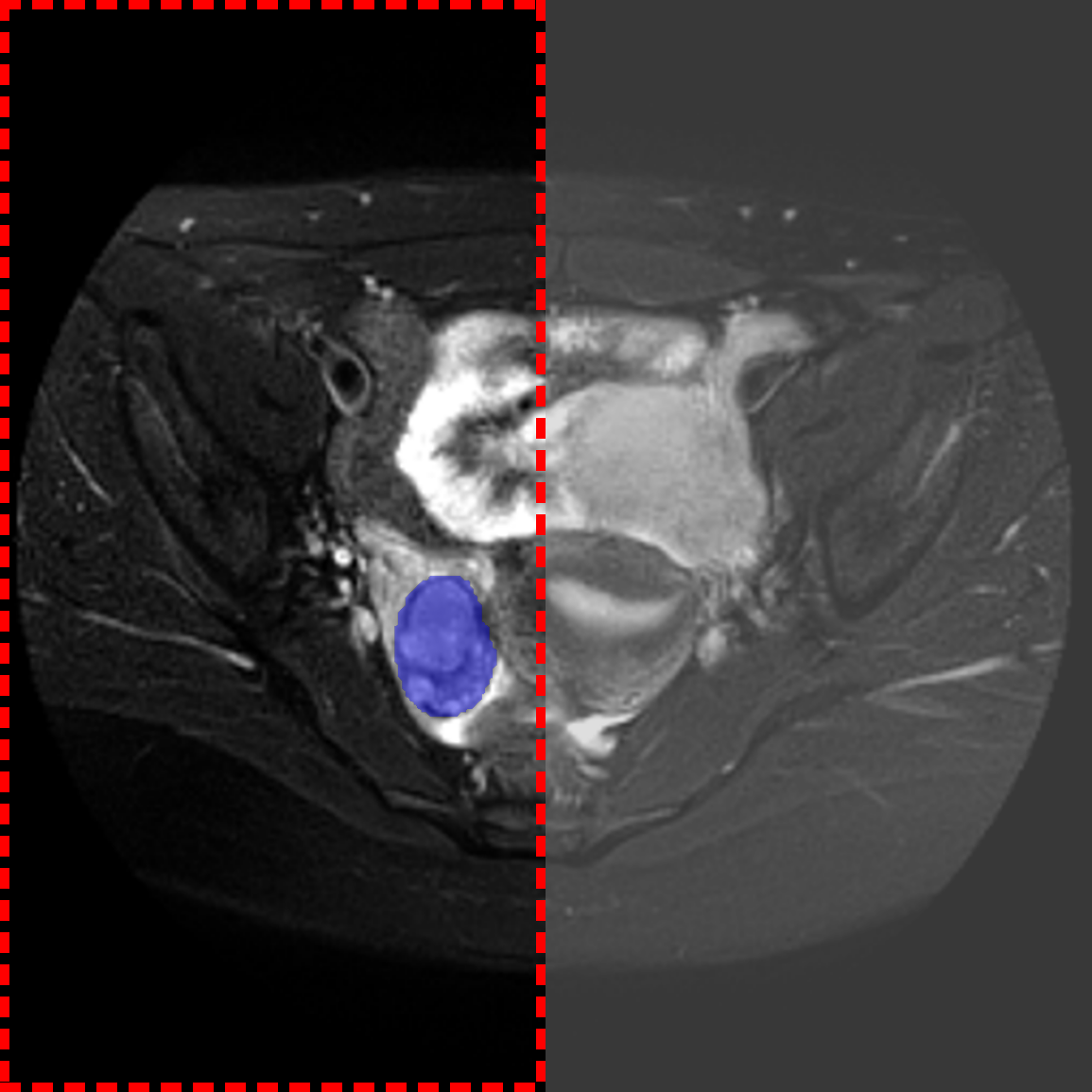} \\

\textbf{D2-061} &
\includegraphics[width=2.05cm]{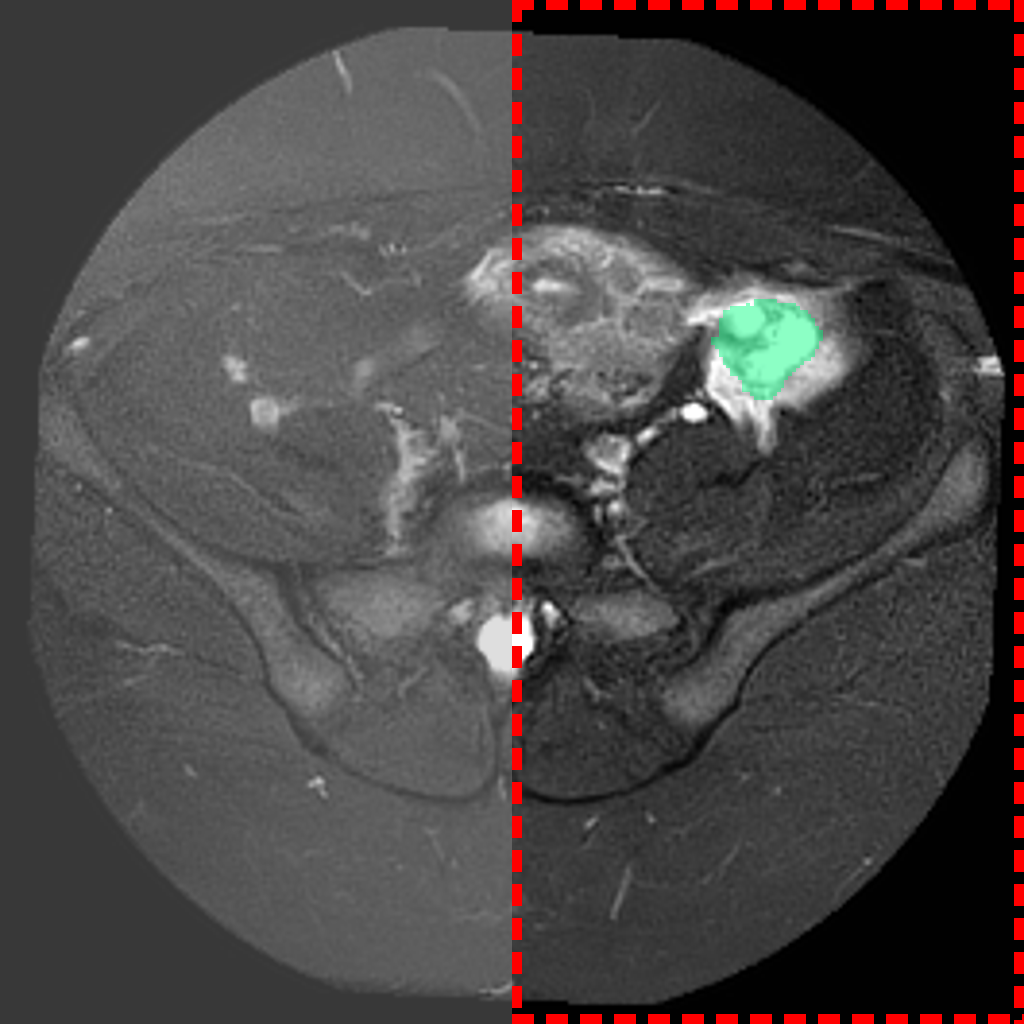} &
\includegraphics[width=2.05cm]{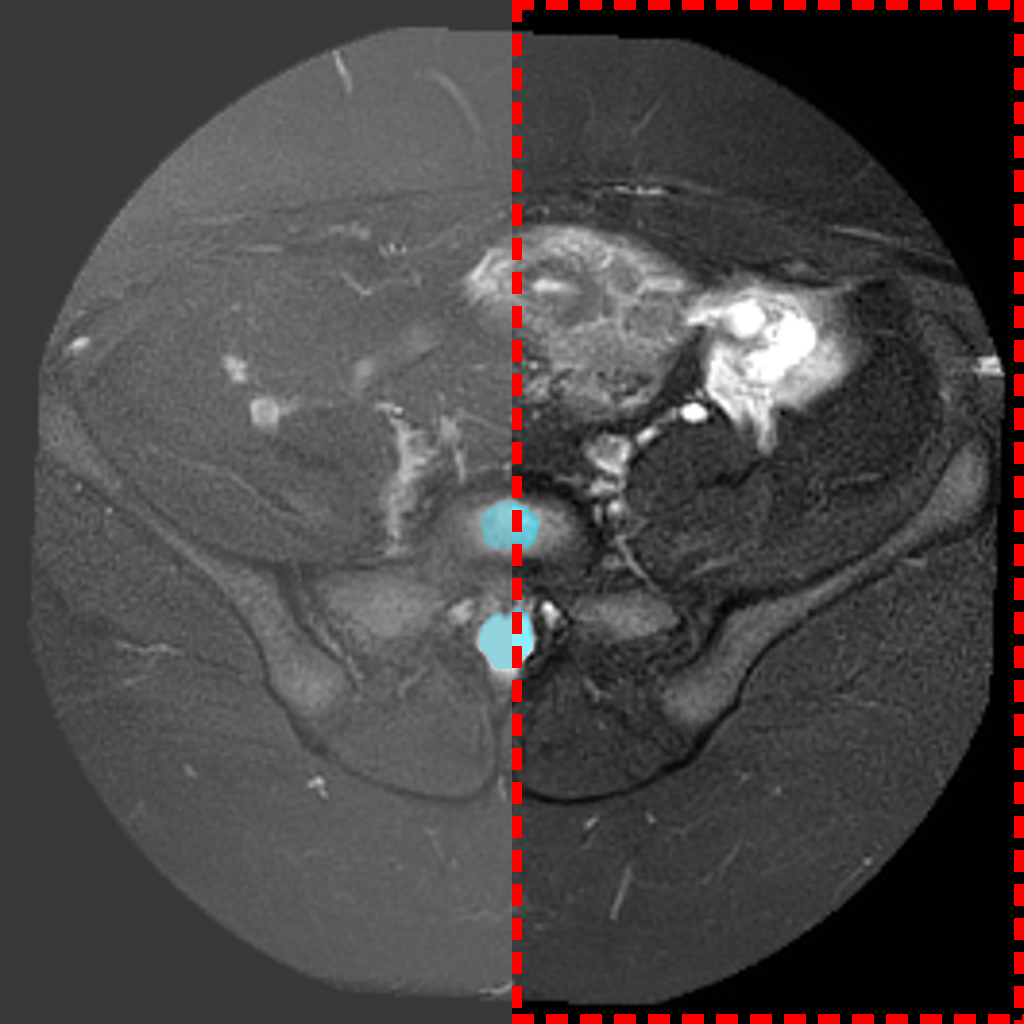} &
\includegraphics[width=2.05cm]{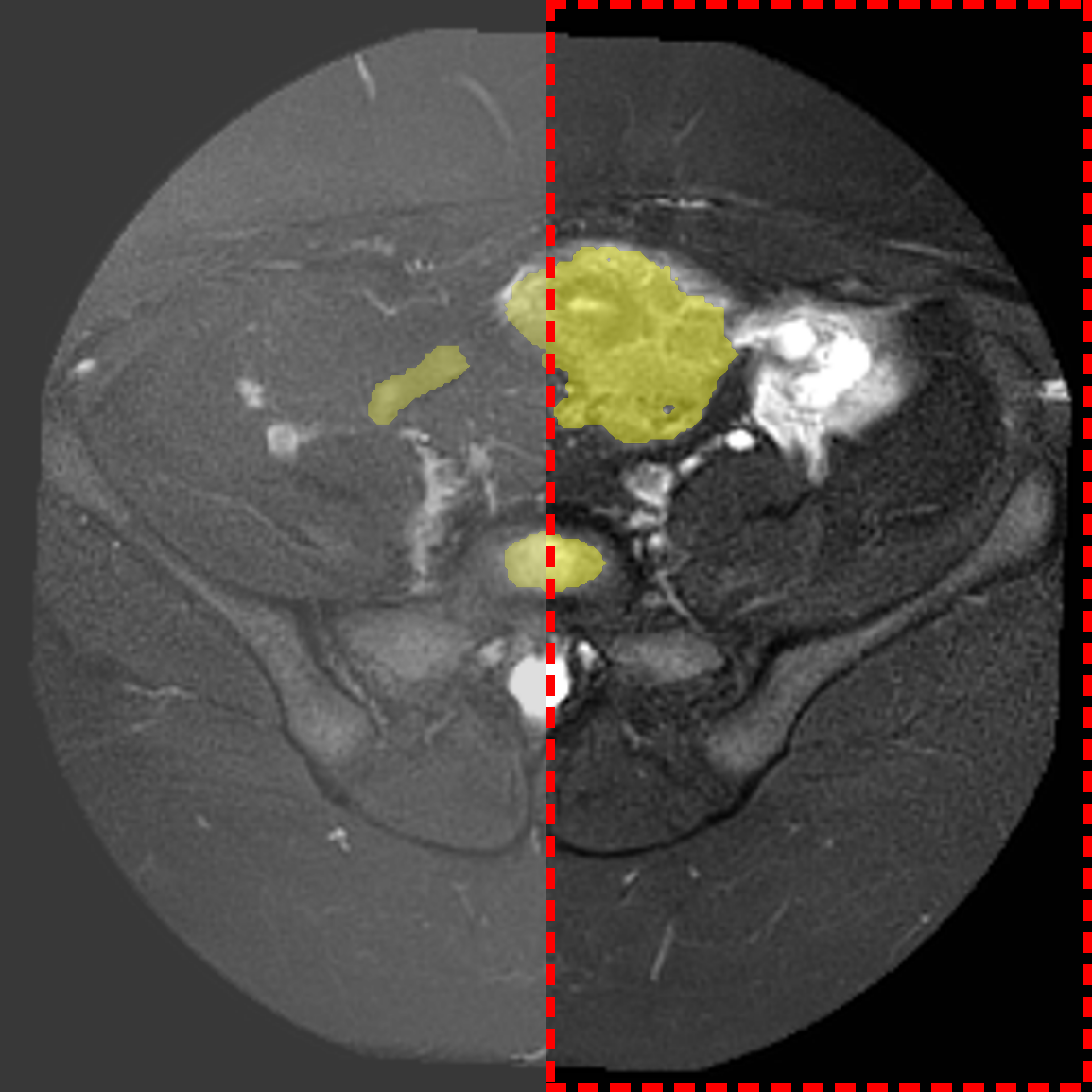} &
\includegraphics[width=2.05cm]{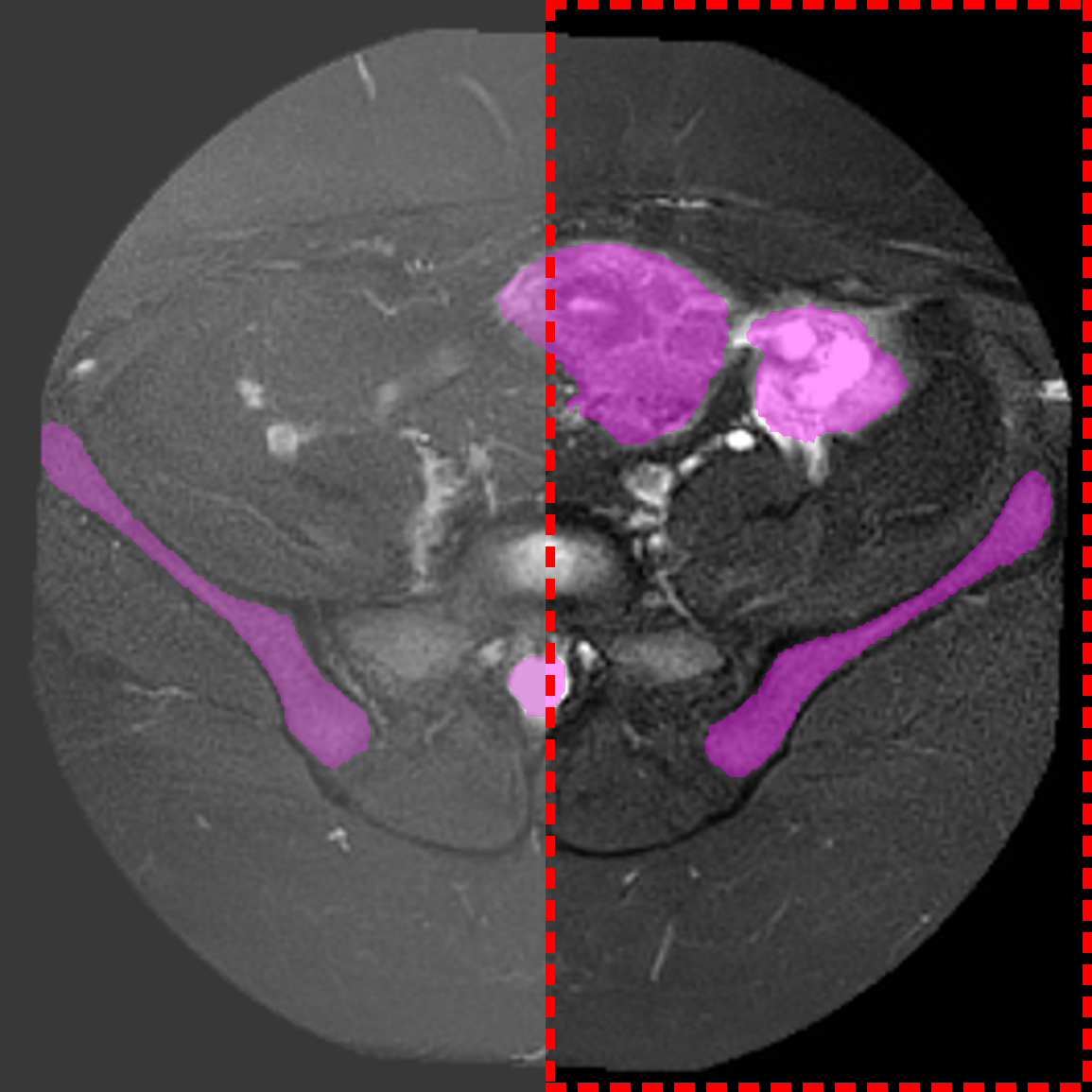} &
\includegraphics[width=2.05cm]{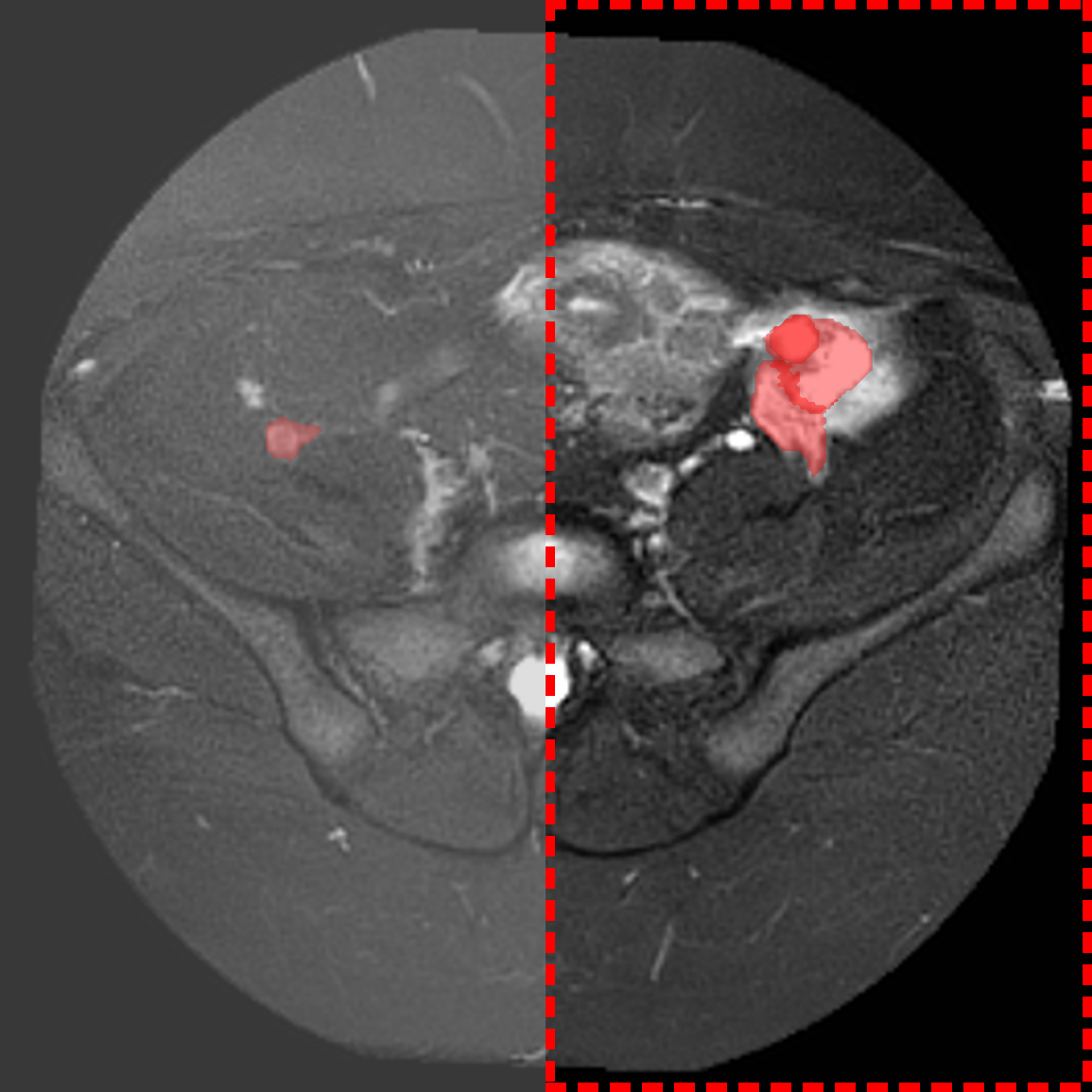} &
\includegraphics[width=2.05cm]{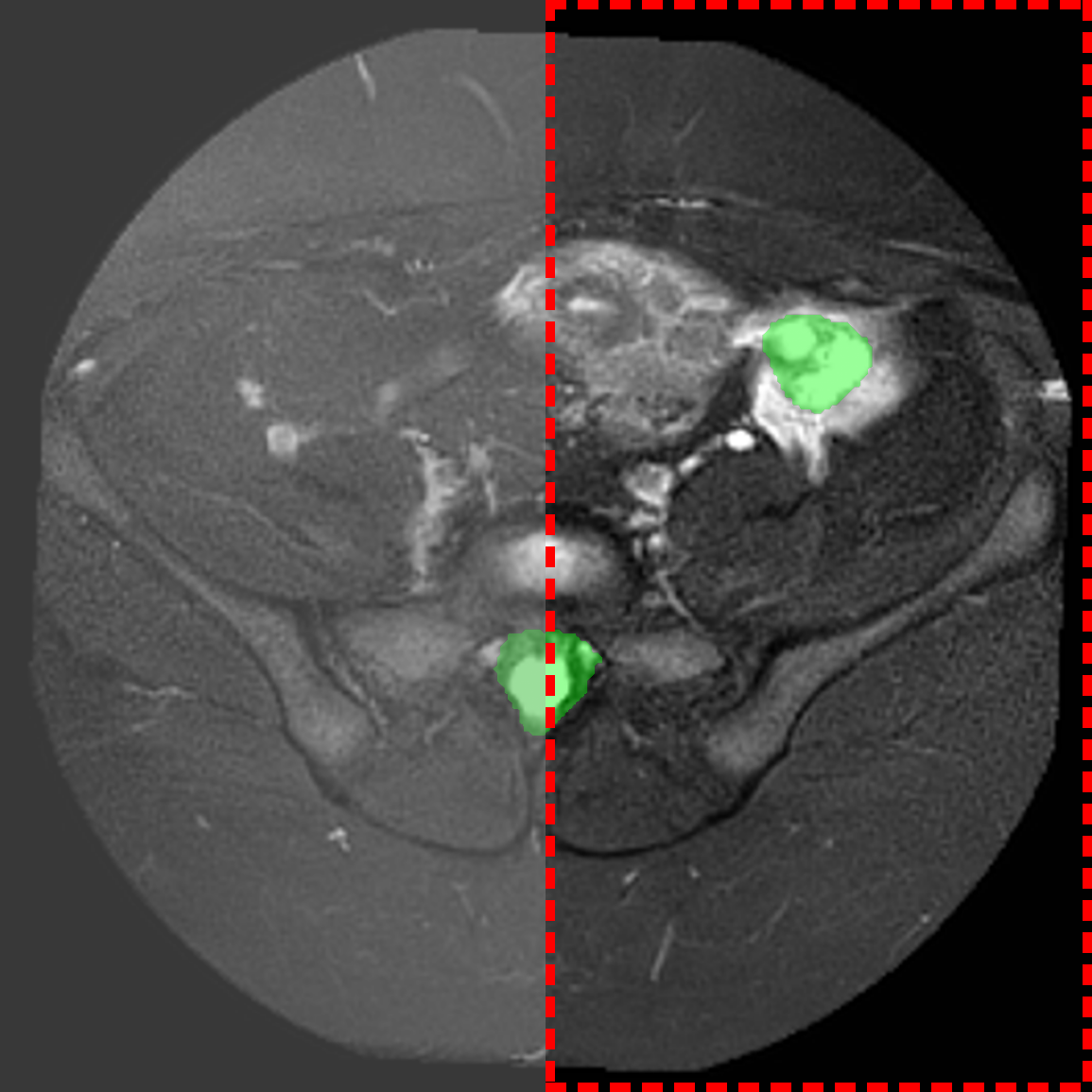} &
\includegraphics[width=2.05cm]{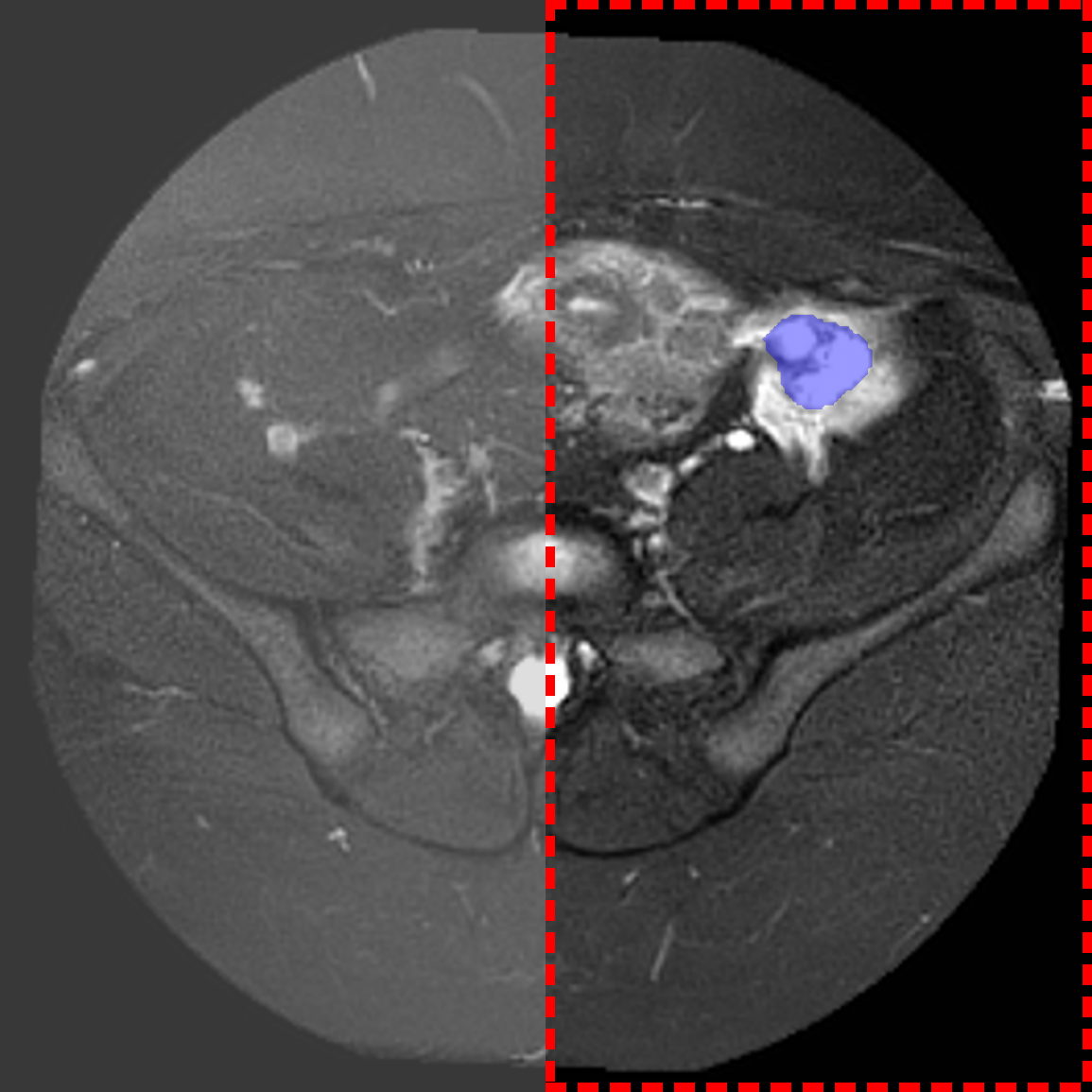}
\end{tabular} %LF: moved table to seperate file for better readability
}
\caption{
\textbf{Qualitative comparison of ovary segmentation results on MRI.}
Each row corresponds to one testing case, and the columns show the ground truth (GT) and segmentation results from different methods. "FT" denotes fine-tuning. 
%The red dashed box denotes the valid evaluation region, defined as the half-plane containing the annotated ovary mask. Model's predictions outside this region were ignored during loss computation and excluded from quantitative metric calculation.
The red dashed box indicates the valid evaluation region, with predictions outside it ignored during training and evaluation.
}
\label{fig:qualitative}
\end{figure}

% qualitative segmentation results from three MRI testing cases are shown in~\cref{fig:qualitative}. 
\Cref{fig:qualitative} shows qualitative segmentation results for three representative MRI test cases.
The nnU-Net baseline struggles to identify the expected ovarian region in the given MRI image, while the zero-shot SAM3 and MedSAM3 models tend to produce large false-negative regions in the pelvic area and show poor capability in localizing small ovarian structures. Fine-tuning improves the overall localization ability, while incorporating the prototype prior further refines the predicted boundary of the ovary regions. Nevertheless, the proposed model still struggles to accurately segment the entire ovarian region. 
% \begin{figure}[t!]
%     \centering
%     \includegraphics[width=0.99\linewidth]{figs/Results.pdf}
%     \caption{\textbf{Qualitative comparison of ovary segmentation results on MRI. }Each row corresponds to one testing case, and the columns show the ground truth (GT), ground-truth mask (GT Mask), and segmentation results from different methods. \textit{w.o. FT} denotes without fine-tuning, and \textit{w. P} denotes with the proposed prototype priors from TVUS. }
%     \label{fig:qualitative}
% \end{figure}

\subsection{Ablation Study}
\label{sec:additional_exp}
% To evaluate the generalizability of the proposed framework, we further conducted experiments using SAM3~\cite{carion2025sam} as the backbone architecture. Compared with the fine-tuned SAM3 baseline, incorporating the proposed prototype-based dual-branch framework resulted in improved segmentation performance (mDice: xxx vs. xxx, mAP: xxx vs. xxx). These results demonstrate that the performance gains are not specific to MedSAM3 and suggest that the proposed prototype bank provides a robust modality-bridging prior that can be effectively integrated with different foundation-model backbones.
We further investigate the contributions of the prototype bank and warm-up strategy through the ablation study in~\cref{tab:quantitative2}. 
%When removing the proposed prototype bank, the negative samples are directly from the background prototypes of the current TVUS image batch. 
%Under this setting, the performance drops from 61.02\% to 57.74\% in terms of mDICE, while the model's object detection ability slightly improves (mAP: 27.28\% vs. 28.37\%). 
Removing the prototype bank and directly using TVUS features in the contrastive loss reduces mDICE from 61.02\% to 57.74\%. In contrast, mAP slightly increases from 27.28\% to 28.37\%, suggesting that the prototype bank primarily improves segmentation quality rather than object localization.
%Surprisingly, jointly warming up the TVUS and MRI branches leads to a clear performance degradation. By contrast, warming up only the TVUS branch for 3 epochs achieves the best performance. However, longer warm-up schedules on TVUS branch (5 or 10 epochs) do not bring further improvements and even reduce the segmentation accuracy. 
Surprisingly, performance drops when both branches are jointly warmed up or when no warm up is used. Moreover, longer TVUS warm-up schedules (5 or 10 epochs) do not yield further improvements and instead reduce segmentation accuracy. Overall, warming up only the TVUS branch for 3 epochs achieves the best performance. 
In addition, we conducted sensitivity analysis on the contrastive loss weight $\lambda$ and EMA momentum $\alpha$ as shown in~\cref{tab:sensitivity}, indicating the selected parameter settings achieve the best performance in segmenting ovaries, while the proposed framework shows a robust performance against parameter changes in general. 

\begin{table}[t!]
    \centering
    \caption{\textbf{Performance comparison about different prototype bank and warm-up setting.} First 2 columns indicate which branch has been activated in the warm-up phase. All of the experiments use MedSAM3 as backbone model. The best performance of each metric is in \textbf{bold}, second best \underline{underlined}.}
    %The MRI branch are all trained with default batch size and training epochs.}
    %\vspace{0.5em}

        % \begin{tabular}{@{}cccccccc@{}}
    %     \toprule

    %       \multirow{2}{*}{\textbf{TVUS}}
    %     & \multirow{2}{*}{\textbf{MRI}}
    %     & \multirow{2}{*}{\textbf{Epoch(s)}}
    %     & \multirow{2}{*}{\makecell[c]{\textbf{With}\\\textbf{Bank}}}
    %     & \multirow{2}{*}{mDICE}
    %     & \multirow{2}{*}{\makecell[c]{mAP\\IoU=0.50:0.95}}
    %     & \multirow{2}{*}{\makecell[c]{mAP\\@50}}
    %     & \multirow{2}{*}{\makecell[c]{mAP\\@75}} \\

    %     % \cmidrule(lr){1-2}
    %     % TVUS & MRI & & & & & & \\
    %     \\

    %     \midrule

    %     \cmark &        & 3  & \cmark & \textbf{61.02$\pm$3.40} & 27.28$\pm$0.98 & 59.28$\pm$3.46 & 19.68$\pm$2.41 \\
    %     \cmark &        & 3  &        & \underline{57.74$\pm$4.19} & \textbf{28.37$\pm$1.52} & \textbf{62.69$\pm$3.27} & \textbf{19.82$\pm$1.50} \\

    %     \midrule

    %     \cmark & \cmark & 3  & \cmark & 17.92$\pm$2.46 & 6.45$\pm$2.25  & 19.28$\pm$6.63 & 2.28$\pm$1.41 \\
    %             \midrule

    %            &        & 0  & \cmark & 45.96$\pm$5.49 & 24.92$\pm$1.58 & 54.99$\pm$4.89 & 15.19$\pm$3.42 \\
    %     \cmark &        & 5  & \cmark & 50.86$\pm$6.70 & 27.63$\pm$0.53 & 61.81$\pm$1.73 & 18.31$\pm$2.83 \\
    %     \cmark &        & 10 & \cmark & 55.11$\pm$5.18 & \underline{28.12$\pm$0.71} & \underline{62.49$\pm$3.01} & \underline{19.73$\pm$1.91} \\

    %     \bottomrule
    % \end{tabular}

\setlength{\tabcolsep}{10pt} % default is 6pt
%\small
\footnotesize
%\scriptsize

        \begin{tabular}{@{}cccccc@{}}
        \toprule
        \multicolumn{3}{c}{\textbf{Warm-up phase}} &
        \multicolumn{1}{c}{\textbf{With}} &
        \multicolumn{2}{c}{} \\
        \cmidrule(lr){1-3} %\cmidrule(lr){5-5}
        \textbf{TVUS} &
        \textbf{MRI} &
        \textbf{Epoch(s)} &
        \textbf{Bank} &
        \textbf{mDice} &
        \textbf{mAP} \\

        % \cmidrule(lr){1-2}
        % TVUS & MRI & & & & & & \\

        \midrule

        \cmark &        & 3  & \cmark & \textbf{61.02$\pm$3.40} & 27.28$\pm$0.98  \\
        \cmark &        & 3  &        & \underline{57.74$\pm$4.19} & \textbf{28.37$\pm$1.52} \\

        \midrule

        \cmark & \cmark & 3  & \cmark & 17.92$\pm$2.46 & 6.45$\pm$2.25 \\
               &        & 0  & \cmark & 45.96$\pm$5.49 & 24.92$\pm$1.58  \\

                        \midrule

        \cmark &        & 5  & \cmark & 50.86$\pm$6.70 & 27.63$\pm$0.53  \\
        \cmark &        & 10 & \cmark & 55.11$\pm$5.18 & \underline{28.12$\pm$0.71} \\

        \bottomrule
    \end{tabular}

    \label{tab:quantitative2}
\end{table}
\begin{table}[t!]
    \centering
    \caption{\textbf{Sensitivity analysis about key parameters.} $\lambda$ indicates the contrastive loss weight, while $\alpha$ denotes the EMA momentum for updating the TVUS prototype. The best performance of each metric is in \textbf{bold}. }
    %\vspace{0.5em}

    \begingroup
\setlength{\tabcolsep}{4pt}
\renewcommand{\arraystretch}{0.9}
\footnotesize

\begin{tabular}{@{}cccccccc@{}}
\toprule
\multicolumn{4}{c}{$\lambda$} &
\multicolumn{4}{c}{$\alpha$} \\
\cmidrule(lr){1-4}\cmidrule(lr){5-8}
Value & mDICE & mAP & & Value & mDICE & mAP & \\
\midrule
0.005 & $58.82\pm5.22$ & $27.19\pm0.84$ &&
0.75 & $57.96\pm11.14$ & $26.61\pm1.64$ &\\

0.010 & $57.23\pm7.99$ & $26.09\pm1.31$ &&
0.80 & $59.42\pm5.36$ & $27.77\pm0.81$ &\\

0.015 & $\mathbf{61.02\pm3.40}$ & $\mathbf{27.28\pm0.98}$ &&
0.85 & $59.56\pm9.03$ & $25.13\pm1.43$ &\\

0.020 & $59.07\pm5.24$ & $26.70\pm1.77$ &&
0.90 & $59.31\pm7.21$ & $\mathbf{28.04\pm1.27}$ &\\

0.025 & $56.51\pm10.37$ & $26.35\pm2.86$ &&
0.95 & $\mathbf{61.02\pm3.40}$ & $27.28\pm0.98$ &\\
\bottomrule
\end{tabular}

\endgroup

    \label{tab:sensitivity}
\end{table}

\section{Discussion and Conclusion}
This study explores whether unpaired TVUS ovary data can provide a useful cross-modal prototype prior for text-promptable MRI ovary segmentation. By leveraging unpaired TVUS ovary masks, the proposed prototype contrastive learning framework constructs a population-level ovary foreground prior without requiring patient-level correspondence or cross-modal registration. Quantitatively, the prototype prior improves mDICE and overall mAP. Qualitatively, the proposed method produces more compact predictions and reduces false-positive responses in surrounding pelvic regions.

Our results show that zero-shot SAM3 and MedSAM3 have limited performance on female pelvic MRI, highlighting the need for target-domain adaptation.
While the proposed dual-branch framework consistently outperforms the fully fine-tuned MedSAM3 baseline in both quantitative metrics and qualitative segmentation quality, the overall performance on MRI ovary segmentation remains modest. A potential reason could be the limited availability of ovary annotations from the original dataset. 
% This may be attributed to the substantial differences between TVUS and MRI. In TVUS images, ovaries typically appear as prominent foreground structures
% with relatively distinctive visual characteristics
% , whereas in MRI they often occupy only a small portion of the image and can be difficult to distinguish from surrounding pelvic organs and lesions. Consequently, the anatomical prototype learned from TVUS may only partially transfer to MRI feature representations. This suggests that the proposed prototype primarily acts as a weak feature-level regularizer, providing complementary anatomical guidance rather than replacing the need for target-domain segmentation supervision.
%
Furthermore, the current background prototypes are extracted from the entire non-ovary region rather than the near-foreground region around the ovary. As a result, the background prototype may be dominated by easy background pixels and %may not sufficiently represent confusing structures adjacent to the ovary. 
fail to adequately represent anatomically similar structures adjacent to the ovary, limiting their ability to provide informative negative supervision during contrastive learning.
%However, the ablation results, indicate that the proposed prototype-based prior contributes positively and robustly to the overall framework.
The ablation study shows that a three-epoch TVUS-only warm-up achieves the best performance, balancing prototype stability and cross-modal feature alignment. However, modality-specific LoRA adaptation may still cause feature drift and weaken the tied projector's ability to learn a coherent joint space. 

Future work will focus on constructing near-background prototypes using dilated ovary masks. Another direction is to explore the prototype-based self-supervised pretraining to improve the feature alignment between TVUS and MRI branch. We also plan to extend the framework to 3D volumetric segmentation and evaluate its generalizability to other modalities and downstream tasks, such as laparoscopic video analysis, uterus segmentation, and cyst detection. 
 %% removed for anonymized MICCAI submission.
    
    % The following acknowledgement and disclaimer sections can be removed for the double-blind review process.  If and when your paper is accepted, reinsert the acknowledgement and the disclaimer clause in your final camera-ready version.
    % IF you opted to include the acknowledgement and disclaimer sections, they will count towards the 8-page limit.

\begin{credits}
\subsubsection{\ackname}This study was supported by the Bavarian State Ministry of Health, Care and Prevention in the context of the project EndoKI. The authors gratefully acknowledge the scientific support and HPC resources provided by the Erlangen National High Performance Computing Center (NHR@FAU) of the Friedrich-Alexander-Universität Erlangen-Nürnberg (FAU). The hardware is partially funded by the German Research Foundation (DFG).

\subsubsection{\discintname}
The authors have no competing interests to declare that are relevant to the content of this article.
\end{credits}

%
% ---- Bibliography ----
%
% BibTeX users should specify bibliography style 'splncs04'.
% References will then be sorted and formatted in the correct style.
%
\bibliographystyle{splncs04}
\bibliography{capi}

@article{olive2001treatment,
  title={Treatment of endometriosis},
  author={Olive, David L and Pritts, Elizabeth A},
  journal={New England Journal of Medicine},
  volume={345},
  number={4},
  pages={266--275},
  year={2001},
  publisher={Mass Medical Soc}
}

@article{mittal2025artificial,
  title={Artificial intelligence applications in endometriosis imaging},
  author={Mittal, Sneha and Tong, Angela and Young, Scott and Jha, Priyanka},
  journal={Abdominal Radiology},
  volume={50},
  number={10},
  pages={4901--4913},
  year={2025},
  publisher={Springer}
}

@article{daniilidis2022transvaginal,
  title={Transvaginal ultrasound in the diagnosis and assessment of endometriosis — {A}n overview: {H}ow, why, and when},
  author={Daniilidis, Angelos and Grigoriadis, Georgios and Dalakoura, Dimitra and D’Alterio, Maurizio N and Angioni, Stefano and Roman, Horace},
  journal={Diagnostics},
  volume={12},
  number={12},
  pages={2912},
  year={2022},
  publisher={MDPI}
}

@article{thomassin2025esur,
  title={{ESUR} consensus {MRI} for endometriosis: {I}ndications, reporting, and classifications},
  author={Thomassin-Naggara, Isabelle and Dolciami, Miriam and Chamie, Luciana P and Guerra, Adalgisa and Bharwani, Nishat and Freeman, Susan and Rousset, Pascal and Manganaro, Lucia},
  journal={European Radiology},
  volume={35},
  number={11},
  pages={7260--7268},
  year={2025},
  publisher={Springer}
}

@article{galczynski2019ovarian,
  title={Ovarian endometrioma - {A} possible finding in adolescent girls and young women: {A} mini-review},
  author={Ga{\l}czy{\'n}ski, Krzysztof and J{\'o}{\'z}wik, Maciej and Lewkowicz, Dorota and Semczuk-Sikora, Anna and Semczuk, Andrzej},
  journal={Journal of Ovarian Research},
  volume={12},
  number={1},
  pages={104},
  year={2019},
  publisher={Springer}
}

@inproceedings{butler2023effectiveness,
  title={The effectiveness of self-supervised pre-training for multi-modal endometriosis classification},
  author={Butler, David and Wang, Hu and Zhang, Yuan and To, Minh-Son and Condous, George and Leonardi, Mathew and Knox, Steven and Avery, Jodie and Hull, M Louise and Carneiro, Gustavo},
  booktitle={2023 45th Annual International Conference of the IEEE Engineering in Medicine \& Biology Society (EMBC)},
  pages={1--5},
  year={2023},
  organization={IEEE}
}

@article{wang2025human,
  title={Human--{AI} collaborative multi-modal multi-rater learning for endometriosis diagnosis},
  author={Wang, Hu and Butler, David and Zhang, Yuan and Avery, Jodie and Knox, Steven and Ma, Congbo and Hull, Louise and Carneiro, Gustavo},
  journal={Physics in Medicine \& Biology},
  volume={70},
  number={1},
  pages={015008},
  year={2025},
  publisher={IOP Publishing}
}

@article{zhang2025unpaired,
  title={Unpaired multi-modal training and single-modal testing for detecting signs of endometriosis},
  author={Zhang, Yuan and Wang, Hu and Butler, David and Smart, Brandon and Xie, Yutong and To, Minh-Son and Knox, Steven and Condous, George and Leonardi, Mathew and Avery, Jodie C and others},
  journal={Computerized Medical Imaging and Graphics},
  volume={124},
  pages={102575},
  year={2025},
  publisher={Elsevier}
}

@inproceedings{zhang2023distilling,
  title={Distilling missing modality knowledge from ultrasound for endometriosis diagnosis with magnetic resonance images},
  author={Zhang, Yuan and Wang, Hu and Butler, David and To, Minh-Son and Avery, Jodie and Hull, M Louise and Carneiro, Gustavo},
  booktitle={2023 IEEE 20th International Symposium on Biomedical Imaging (ISBI)},
  pages={1--5},
  year={2023},
  organization={IEEE}
}

@article{tank2025automatic,
  title={Automatic uterus segmentation in transvaginal ultrasound using {U-Net} and {nnU-Net}},
  author={Tank, Dilara and Schor, Bianca GS and Trommelen, Lisa M and Huirne, Judith AF and Calixto, Iacer and de Leeuw, Robert A},
  journal={PLoS One},
  volume={20},
  number={11},
  pages={e0336237},
  year={2025},
  publisher={Public Library of Science San Francisco, CA USA}
}

@article{lyu2025unsupervised,
  title={Unsupervised cross-domain semantic segmentation on multi-modality ovarian tumor ultrasound data},
  author={Lyu, Shuchang and Zhao, Qi and Bai, Wenpei and Cai, Linghan and Cheng, Guangliang and Cui, Guangxia and Yang, Min and Chen, Lijiang and Zhou, Huiyu},
  journal={Pattern Recognition},
  pages={112311},
  year={2025},
  publisher={Elsevier}
}

@article{zhao2022mmotu,
  title={{MMOTU}: {A} multi-modality ovarian tumor ultrasound image dataset for unsupervised cross-domain semantic segmentation},
  author={Zhao, Qi and Lyu, Shuchang and Bai, Wenpei and Cai, Linghan and Liu, Binghao and Cheng, Guangliang and Wu, Meijing and Sang, Xiubo and Yang, Min and Chen, Lijiang},
  journal={arXiv preprint arXiv:2207.06799},
  year={2022}
}

@article{bonevs2024automatic,
  title={Automatic segmentation and alignment of uterine shapes from {3D} ultrasound data},
  author={Bone{\v{s}}, Eva and Gergolet, Marco and Bohak, Ciril and Lesar, {\v{Z}}iga and Marolt, Matija},
  journal={Computers in Biology and Medicine},
  volume={178},
  pages={108794},
  year={2024},
  publisher={Elsevier}
}

@article{liang2025multi,
  title={A Multi-Modal Pelvic {MRI} Dataset for Deep Learning-Based Pelvic Organ Segmentation in Endometriosis},
  author={Liang, Xiaomin and Alpuing Radilla, Linda A and Khalaj, Kamand and Dawoodally, Haaniya and Mokashi, Chinmay and Guan, Xiaoming and Roberts, Kirk E and Sheth, Sunil A and Tammisetti, Varaha S and Giancardo, Luca},
  journal={Scientific Data},
  volume={12},
  number={1},
  pages={1292},
  year={2025},
  publisher={Nature Publishing Group UK London}
}

@article{isensee2021nnu,
  title={{nnU-Net}: {A} self-configuring method for deep learning-based biomedical image segmentation},
  author={Isensee, Fabian and Jaeger, Paul F and Kohl, Simon AA and Petersen, Jens and Maier-Hein, Klaus H},
  journal={Nature Methods},
  volume={18},
  number={2},
  pages={203--211},
  year={2021},
  publisher={Nature Publishing Group US New York}
}

@article{arjomandi2026prompts,
  title={From Prompts to Pipelines: {E}valuating {LLM}-Generated Medical Image Segmentation Baselines},
  author={Arjomandi, Jasmin and Neubig, Luisa and Mathis-Ullrich, Franziska and Kist, Andreas M and others},
  journal={Machine Learning for Biomedical Imaging},
  volume={2026},
  number={MELBA--BVM 2025 Special Issue},
  pages={159--183},
  year={2026}
}

@article{liu2025medsam3,
  title={{MedSAM3}: {D}elving into Segment Anything with Medical Concepts},
  author={Liu, Anglin and Xue, Rundong and Cao, Xu R and Shen, Yifan and Lu, Yi and Li, Xiang and Chen, Qianqian and Chen, Jintai},
  journal={arXiv preprint arXiv:2511.19046},
  year={2025}
}

@article{han2024parameter,
  title={Parameter-efficient fine-tuning for large models: {A} comprehensive survey},
  author={Han, Zeyu and Gao, Chao and Liu, Jinyang and Zhang, Jeff and Zhang, Sai Qian},
  journal={arXiv preprint arXiv:2403.14608},
  year={2024}
}

@article{carion2025sam,
  title={{SAM 3}: {S}egment anything with concepts},
  author={Carion, Nicolas and Gustafson, Laura and Hu, Yuan-Ting and Debnath, Shoubhik and Hu, Ronghang and Suris, Didac and Ryali, Chaitanya and Alwala, Kalyan Vasudev and Khedr, Haitham and Huang, Andrew and others},
  journal={arXiv preprint arXiv:2511.16719},
  year={2025}
}

@article{figueredo2024automatic,
  title={Automatic segmentation of deep endometriosis in the rectosigmoid using deep learning},
  author={Figueredo, Weslley Kelson Ribeiro and Silva, Arist{\'o}fanes Corr{\^e}a and de Paiva, Anselmo Cardoso and Diniz, Jo{\~a}o Ot{\'a}vio Bandeira and Brandao, Alice and Oliveira, Marco Aurelio Pinho},
  journal={Image and Vision Computing},
  volume={151},
  pages={105261},
  year={2024},
  publisher={Elsevier}
}

@article{oord2018representation,
  title={Representation learning with contrastive predictive coding},
  author={Oord, Aaron van den and Li, Yazhe and Vinyals, Oriol},
  journal={arXiv preprint arXiv:1807.03748},
  year={2018}
}

@article{saleem2025deep,
  title={Deep Learning-Based Automated Segmentation of Uterine Myomas},
  author={Saleem, Tausifa Jan and Yaqub, Mohammad},
  journal={arXiv preprint arXiv:2508.11010},
  year={2025}
}
%
% \begin{thebibliography}{8}
% \bibitem{ref_article1}
% Author, F.: Article title. Journal \textbf{2}(5), 99--110 (2016)

% \bibitem{ref_lncs1}
% Author, F., Author, S.: Title of a proceedings paper. In: Editor,
% F., Editor, S. (eds.) CONFERENCE 2016, LNCS, vol. 9999, pp. 1--13.
% Springer, Heidelberg (2016). \doi{10.10007/1234567890}

% \bibitem{ref_book1}
% Author, F., Author, S., Author, T.: Book title. 2nd edn. Publisher,
% Location (1999)

% \bibitem{ref_proc1}
% Author, A.-B.: Contribution title. In: 9th International Proceedings
% on Proceedings, pp. 1--2. Publisher, Location (2010)

% \bibitem{ref_url1}
% LNCS Homepage, \url{http://www.springer.com/lncs}, last accessed 2023/10/25
% \end{thebibliography}
\end{document}